\documentclass{article}

\usepackage{microtype}
\usepackage{graphicx}
\usepackage{subfigure}
\usepackage{booktabs} 
\usepackage{enumitem}

\usepackage{hyperref}

\usepackage[accepted]{icml2025}

\usepackage{amsmath}
\usepackage{amssymb}
\usepackage{mathtools}
\usepackage{amsthm}

\usepackage{url}
\usepackage{multirow}
\usepackage{listings}
\usepackage{color}

\usepackage[capitalize,noabbrev]{cleveref}

\theoremstyle{plain}
\newtheorem{theorem}{Theorem}[section]
\newtheorem{proposition}[theorem]{Proposition}

\theoremstyle{definition}

\theoremstyle{remark}

\usepackage[textsize=tiny]{todonotes}

\icmltitlerunning{KBQA-o1: Agentic Knowledge Base Question Answering with Monte Carlo Tree Search}

\begin{document}

\twocolumn[
\icmltitle{KBQA-o1: Agentic Knowledge Base Question Answering\protect\\with Monte Carlo Tree Search}



\icmlsetsymbol{equal}{*}

\begin{icmlauthorlist}
\icmlauthor{Haoran Luo}{bupt,ntu}
\icmlauthor{Haihong E}{bupt}
\icmlauthor{Yikai Guo}{hk2y}
\icmlauthor{Qika Lin}{nus}
\icmlauthor{Xiaobao Wu}{ntu}
\icmlauthor{Xinyu Mu}{bupt}
\icmlauthor{Wenhao Liu}{bupt}
\icmlauthor{Meina Song}{bupt}
\icmlauthor{Yifan Zhu}{bupt}
\icmlauthor{Luu Anh Tuan}{ntu}
\end{icmlauthorlist}

\icmlaffiliation{bupt}{Beijing University of Posts and Telecommunications}
\icmlaffiliation{ntu}{Nanyang Technological University}
\icmlaffiliation{hk2y}{Beijing Institute of Computer Technology and Application}
\icmlaffiliation{nus}{National University of Singapore}

\icmlcorrespondingauthor{Haihong E}{ehaihong@bupt.edu.cn}

\icmlkeywords{Machine Learning, ICML} 

\vskip 0.3in
]



\printAffiliationsAndNotice{} 

\begin{abstract}
Knowledge Base Question Answering (KBQA) aims to answer natural language questions with a large-scale structured knowledge base (KB). Despite advancements with large language models (LLMs), KBQA still faces challenges in weak KB awareness, imbalance between effectiveness and efficiency, and high reliance on annotated data. To address these challenges, we propose KBQA-o1, a novel agentic KBQA method with Monte Carlo Tree Search (MCTS). It introduces a ReAct-based agent process for stepwise logical form generation with KB environment exploration. Moreover, it employs MCTS, a heuristic search method driven by policy and reward models, to balance agentic exploration's performance and search space. With heuristic exploration, KBQA-o1 generates high-quality annotations for further improvement by incremental fine-tuning. Experimental results show that KBQA-o1 outperforms previous low-resource KBQA methods with limited annotated data, boosting Llama-3.1-8B model's GrailQA F1 performance to 78.5\% compared to 48.5\% of the previous sota method with GPT-3.5-turbo. Our code is publicly available at \url{https://github.com/LHRLAB/KBQA-o1}.
\end{abstract}

\section{Introduction}
\label{Section1}

\begin{figure}[t]
\centering
\includegraphics[width=8.1cm]{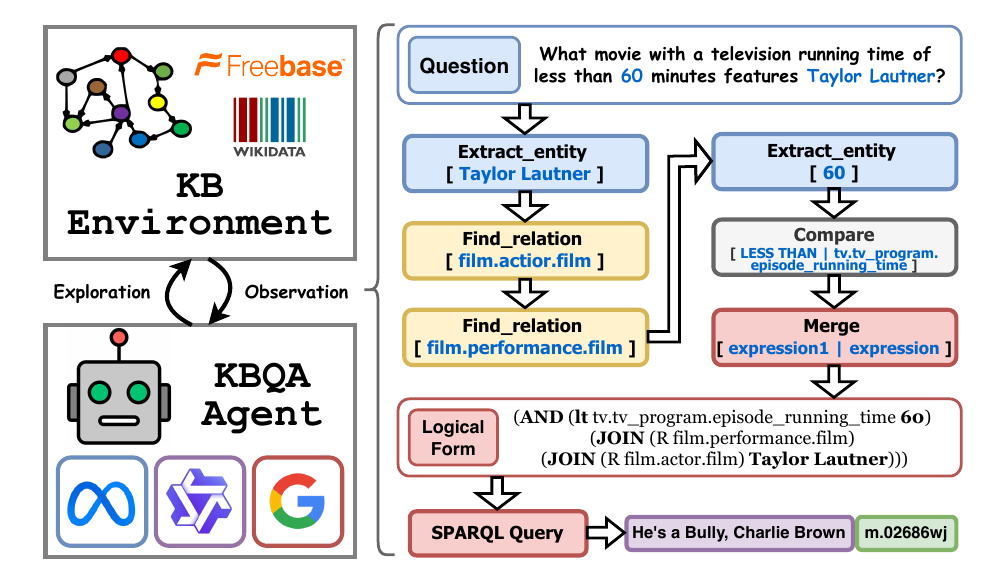}
\caption{
An example of KBQA task to answer a natural language question by exploring the KB environment with KBQA agent.
}
\label{f1}
\end{figure}

Knowledge Base Question Answering (KBQA) leverages a large-scale structured knowledge base (KB), such as Freebase~\cite{Freebase} or Wikidata~\cite{Wikidata}, as a reference to answer questions in natural language, widely applied in fields such as search engines~\cite{searchengine}, medical consultations~\cite{MedGraphRAG}, and legal analysis~\cite{ChatLaw}. Typically, KBs are stored in graph databases and accessed using graph queries, such as SPARQL~\cite{SPARQL}, which support multi-hop and logical queries to acquire knowledge information. To answer natural language questions, language models are usually leveraged~\cite{RnG-KBQA} to convert the questions into logical forms, such as S-expression~\cite{GrailQA}, and then transform them into executable graph queries to obtain answers from KB, as shown in Figure~\ref{f1}.


\begin{figure*}[t]
\centering
\includegraphics[width=0.93\linewidth]{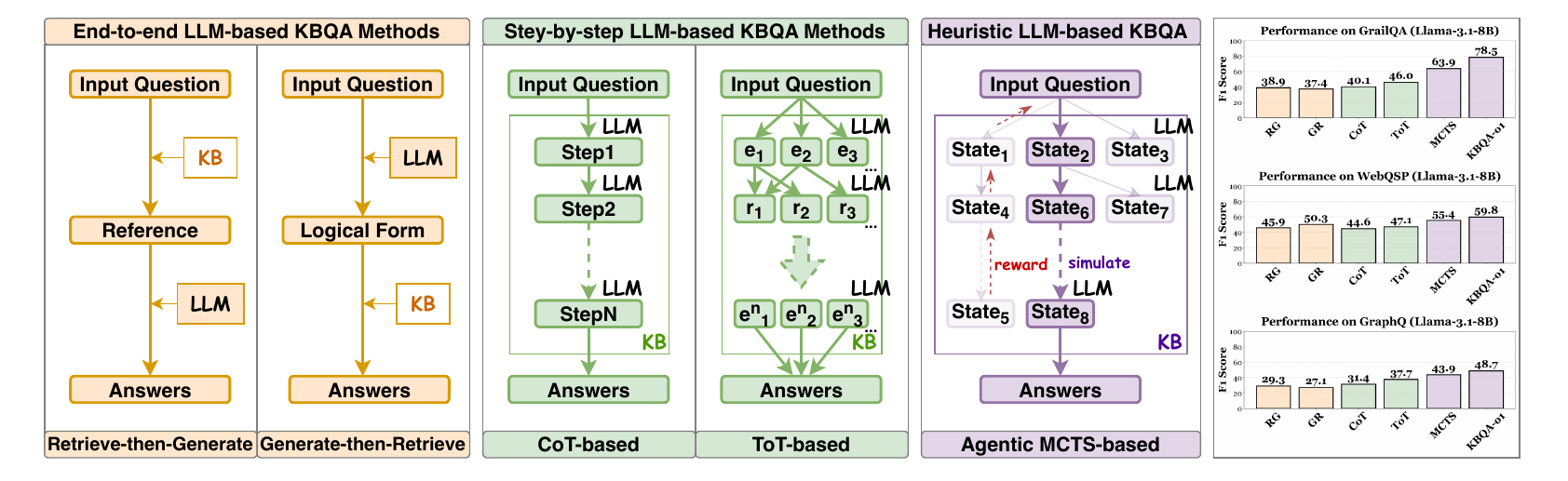}
\caption{Comparison of the previous end-to-end KBQA framework, including retrieve-then-generate (RG) and generate-then-retrieve (GR) methods, step-by-step KBQA methods, including CoT-based and ToT-based methods, and our proposed heuristic KBQA method, which is agentic MCTS-based. With the same Llama-3.1-8B~\cite{llama3} as the base model, both the MCTS-based agent process and the full KBQA-o1 after the incremental fine-tuning show improvements on all three KBQA datasets.}
\label{f2}
\end{figure*}

With the emergence of LLMs~\citep{GPT4,llama3}, two main types of KBQA methods appear, as shown in \Cref{f2}. On the one hand, end-to-end methods~\citep{RoG,ChatKBQA} generate logical forms directly from natural language questions and utilize retrieval before or after generation for improvement.
On the other hand, step-by-step methods~\citep{QueryAgent,ToG} alternate between generation and retrieval for stepwise thinking on KB, performing in a Chain-of-Thought (CoT)~\cite{CoT} or Tree-of-Thoughts (ToT)~\cite{ToT} manner.

However, three main challenges remain:
\textbf{(1) Poor awareness of the KB environment in end-to-end methods.}
Relying on direct logical form generation by language models, end-to-end KBQA methods~\cite{KB-BINDER,KB-Coder} struggle with limited logical form schemas and unseen entities and relations that existed in KB, making it difficult to fully capture the KB environment.
\textbf{(2) Local optima or large search space in step-by-step methods.}
Compared to end-to-end methods, CoT-based step-by-step methods~\cite{QueryAgent} in the search process can lead to local optima due to intermediate bias.
Although ToT-based methods~\cite{ToG} expand search options, tree-like searches still face challenges with large search spaces.
\textbf{(3) Training depends on high-quality annotation.} 
Training open-source LLMs~\cite{ChatKBQA} significantly contributes to the generation of logical forms for KBQA tasks, but this heavily relies on the quality of the annotated data.
For large-scale KBs, producing substantial, high-quality annotated data by human labor is impractical.

To address these challenges, we propose KBQA-o1, a novel agentic KBQA approach with heuristic exploration of the KB environment, as shown in Figure~\ref{f1}. 
First, we design a ReAct-based~\cite{ReAct} agent process with atomic query tools to generate logical forms by fully interacting with the KB environment.
Moreover, we employ Monte Carlo Tree Search (MCTS)~\cite{MCTS}, driven by policy and reward models, to optimize the agent process from local optima or large search space. 
Furthermore, to mitigate reliance on extensive human annotation, we first fine-tune the policy and reward models only using a small amount of sample-labeled data. For unlabeled questions, we employ MCTS exploration with the reward model filter instead of human labor to generate abundant, auto-annotated data, further enhancing the capabilities of both the policy and reward models by incremental fine-tuning.

We perform experiments on three KBQA datasets, GrailQA~\cite{GrailQA}, WebQSP~\cite{WebQSP} and GraphQ~\cite{GraphQ} in low-resource settings~\cite{KB-BINDER} for application with limited annotated data.
Experimental results demonstrate that KBQA-o1 outperforms existing low-resource KBQA methods and even approaches or surpasses the performance of fully supervised KBQA models, especially in more difficult cases like compositional and zero-shot. 
Ablation studies further validate the proposed MCTS-based agent process and incremental fine-tuning, both of which make KBQA-o1 outperform other forms of KBQA methods, as shown in Figure~\ref{f2}. 
In addition, KBQA-o1 supports multiple open-source LLMs, including Llama-3~\cite{llama3}, Qwen2.5~\cite{Qwen2.5} and Gemma-2~\cite{Gemma2}, making it a plug-and-play and promising solution for diverse KBQA applications.

\begin{table*}[t]
\caption{\label{t1}
Description of proposed atomic query tools and corresponding arguments, where $\mathcal{M}_o=\{max, min\}$, and $\mathcal{M}_c=\{<,\leq,>,\geq\}$ are mode sets of Order and Compare. By using tools, the agent can obtain the target functions and equivalent logical forms accordingly.
}
\fontsize{8pt}{8pt}\selectfont
\centering
\vskip 0.1in
\renewcommand\arraystretch{1.1}
\resizebox{\linewidth}{!}{
\begin{tabular}{llll}
\toprule
\textbf{Atomic Query Tool} & \textbf{Arguments} & \textbf{Target Function} & \textbf{Equivalent Logical Form}  \\ \midrule
\texttt{Extract\_entity} {[}\ \textit{entity}\ {]} & \textit{entity} $\in \mathcal{E}$ & \textit{expression} = START(`\textit{entity}') & \textit{entity} \\
\texttt{Find\_relation} {[}\ \textit{relation}\ {]} & \textit{relaiton} $\in \mathcal{R}$ & \textit{expression} = JOIN(`\textit{relation}', \textit{expression}) & (JOIN \textit{relation} (\textit{expression})) \\
\texttt{Merge} {[}\ \textit{expression1}\ $\vert$\ \textit{expression}\ {]}  & \textit{expression1}, \textit{expression} & \textit{expression} = AND(\textit{expression1}, \textit{expression}) & (AND (\textit{expression1}) (\textit{expression})) \\
\texttt{Order} {[}\ \textit{mode}\ $\vert$\ \textit{relation}\ {]} & \textit{mode} $\in \mathcal{M}_o$, \textit{relation} $\in \mathcal{R}$ & \textit{expression} = ARG(`\textit{mode}', \textit{expression}, `\textit{relation}') & (\textit{mode} (\textit{expression}) \textit{relation}) \\
\texttt{Compare} {[}\ \textit{mode}\ $\vert$\ \textit{relation}\ {]} & \textit{mode} $\in \mathcal{M}_c$, \textit{relation} $\in \mathcal{R}$ & \textit{expression} = CMP(`\textit{mode}', `\textit{relation}', \textit{expression}) & (\textit{mode} \textit{relation} (\textit{expression})) \\
\texttt{Time\_constraint} {[}\ \textit{relation}\ $\vert$\ \textit{time}\ {]} & \textit{relation} $\in \mathcal{R}$, \textit{time} $\in \mathcal{E}$ & \textit{expression} = TC(\textit{expression}, `\textit{relation}', `\textit{time}') & (TC (\textit{expression}) \textit{relation} \textit{time}) \\
\texttt{Count} {[}\ \textit{expression}\ {]} & \textit{expression} & \textit{expression} = COUNT(\textit{expression})  & (COUNT (\textit{expression})) \\
\texttt{Finish} {[}\ \textit{expression}\ {]} & \textit{expression} & \textit{expression} = STOP(\textit{expression})  & (\textit{expression}) \\\bottomrule
\end{tabular}
}
\end{table*}

\section{Related Work}
\textbf{Knowledge Base Question Answering.} 
Before the rise of LLMs, KBQA methods could be divided into information-retrieval-based (IR-based)~\citep{GRAFT-Net, PullNet, SR} and semantic-parsing-based (SP-based)~\citep{RnG-KBQA, TIARA}. In the era of LLMs, LLM-based KBQA methods can be divided into two categories: end-to-end and step-by-step. End-to-end methods take advantage of in-context learning (ICL)~\citep{KB-BINDER,KB-Coder} or fine-tuning~\citep{RoG,ChatKBQA} to enable LLMs to generate queries. The step-by-step methods~\citep{Pangu, ToG, QueryAgent} follow a reasoning process on the graph to gradually find the answers. In this paper, we propose the first heuristic KBQA method.

\textbf{LLMs and LLM-powered Agent. }
LLMs have shown significant advantages in generation and reasoning~\cite{llmkg}. Powered by CoT\citep{CoT,CoTpower}, guiding LLMs to think step by step can further enhance reasoning capabilities. ReAct~\citep{ReAct} introduces a prompt-based agent, using tools to interact with the environment. \citet{prm} verifies CoT by rewarding the process and outcome. On the other hand, to expand the thought space in every step, ToT~\citep{ToT} reasons in a tree-like manner. RAP~\citep{RAP} employs MCTS, a heuristic algorithm applied in AlphaGo.

\section{Preliminaries}
\label{Section3}

\textbf{Definition 1: Knowledge Base. }
A knowledge base (KB) is a large-scale knowledge graph $\mathcal{G}=(\mathcal{E},\mathcal{R},\mathcal{T})$, composed of an entity set $\mathcal{E}$, a relation set $\mathcal{R}$, and a triple set $\mathcal{T}$. Relations in the relation set $r \in \mathcal{R}$ are used to connect two entities. Each triple $(s,r,o) \in \mathcal{T}$ is formed by (entity, relation, entity), thus $\mathcal{T}=\{(s,r,o)|s\in\mathcal{E},r\in\mathcal{R},o\in\mathcal{E}\}$.

\textbf{Definition 2: Logical Form. }
The logical form $F$ is a multi-hop expression that can convert equally to a graph query $q=\text{Convert}(F)$. 
Each logical form can be divided into a stepwise list of functions $F=[f_i]_{i=1}^l$ with $l$ steps. 

\textbf{Problem Statement. }
In the KBQA task, given a natural language question $\mathcal{Q}$ and a KB $\mathcal{G}$, the goal is to first convert $\mathcal{Q}$ into a logical form $F$, and then an executable graph query $q$. Once executed, the result $\mathcal{A}=\text{Exec}(q,\mathcal{G})$ is a set of entities in the KB $\mathcal{A}\subset\mathcal{E}$ that answer the question $\mathcal{Q}$.

\begin{figure*}[t]
\centering
\includegraphics[width=0.86\linewidth]{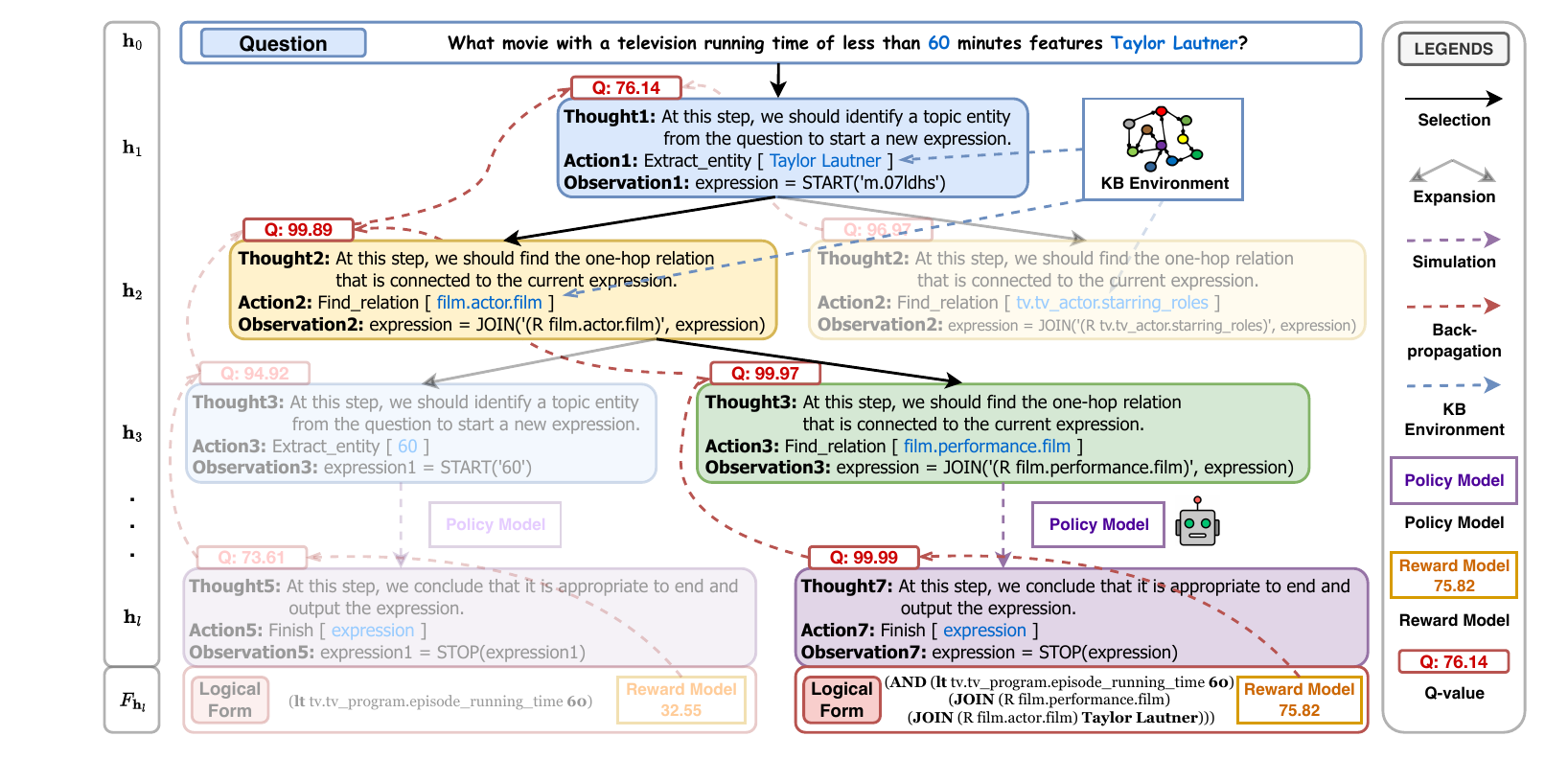}
\caption{An example of the heuristic KB environment exploration with MCTS driven by policy and reward models.}
\label{f3}
\end{figure*}

\section{Method: KBQA-o1}
In this section, we introduce the three components of KBQA-o1: agent initialization, heuristic environment exploration using MCTS with policy and reward models to optimize the agent process, and incremental fine-tuning with auto-annotated data to improve low-resource performance.

\subsection{Agent Initialization}
\label{Section3.1}
KBQA-o1 follows a ReAct-based~\cite{ReAct} agent prompt, with KB environment, the agent state space and exploration space, targeting to generate logical forms.

\textbf{KB Environment} $\mathcal{G}$\textbf{.}
We consider KB a critical environment that provides guides in generating the logical form at each step of the agent, for example, by providing the candidate relations connected to the current state of the logical form. 

\textbf{The Agent State Space} $\mathcal{H}$\textbf{.} 
The agent state $\mathbf{h}_t\ \in \mathcal{H}$ is defined by its exploration history as $\mathbf{h}_t = (\mathbf{h}_0, \mathbf{e}_1, \dots, \mathbf{e}_t)$.
\begin{itemize}
    \item \textbf{Initial State (}$\mathbf{h}_0$\textbf{):} We create an initial prompt, consisting of the task description and the given question $\mathcal{Q}$, as the initial state $\mathbf{h}_0$, as shown in Appendix~\ref{prompt1}.
    \item \textbf{State Update (}$\mathbf{h}_t$\textbf{):} At each step $t$, the state incorporates the latest exploration step, which is prompted as ReAct-based~\cite{ReAct} Thought-Action-Observation tuple $\mathbf{e}_t = (\mathbf{e}_t^{\text{tht}}, \mathbf{e}_t^{\text{act}}, \mathbf{e}_t^{\text{obs}})$ as shown in Appendix~\ref{prompt2}, to update: $\mathbf{h}_t=\mathbf{h}_{t-1}+\mathbf{e}_t$.
    \item \textbf{State Representation (}$F_{\mathbf{h}_t}$\textbf{):} Besides prompt-based history, the cumulative Observations in the trajectory $(\mathbf{e}_1^{\text{obs}}, \dots, \mathbf{e}_t^{\text{obs}})$ determine the function list, which is equivalent to the logical form $F_{\mathbf{h}_t} = [\mathbf{e}_i^{\text{obs}}]_{i=1}^t$.
\end{itemize}

\textbf{The Agent Exploration Space} $\text{Exp}(\mathcal{H},\mathcal{G})$\textbf{.} 
The agent exploration $\mathbf{e}_t \in \text{Exp}(\mathbf{h}_{t-1},\mathcal{G})$ is dynamically determined by the last state $\mathbf{h}_{t-1} \in \mathcal{H}$ and the KB environment $\mathcal{G}$. Each exploration comprises the following components:
\begin{itemize}
    \item \textbf{Tool Selection (}$\mathbf{e}_t^{\text{tht}}$\textbf{):} Based on $\mathbf{h}_{t-1}$, the agent selects one of the eight atomic query tools as shown in \Cref{t1}.
    \item \textbf{Argument Determination (}$\mathbf{e}_t^{\text{act}}$\textbf{):} The agent identifies the appropriate arguments for the selected tool, leveraging the candidates provided by the KB environment. For example:
    \begin{itemize}
        \item Calling \texttt{Extract\_entity} requires specifying an \textit{entity} name existed in $\mathcal{G}$ as the argument.
        \item Calling \texttt{Find\_relation} involves specifying a \textit{relation} name that has a connection with $F_{\mathbf{h}_{t-1}}$.
    \end{itemize}
    \item \textbf{Function Writen (}$\mathbf{e}_t^{\text{obs}}$\textbf{):} Based on the selected tool and arguments, the agent writes down the corresponding target function in Observation, referred to Table~\ref{t1}.
\end{itemize}

\textbf{The Agent Target} $(\mathbf{h}_l,F_{\mathbf{h}_l},\mathcal{A}_{\mathbf{h}_l})$\textbf{.} 
The agent explores the exploration space until calling the \texttt{Finish} tool or when the length of the function list exceeds the maximum allowable length $l<L$. The ultimate target of the agent is to find a complete state $\mathbf{h}_l$, which forms a logical form $F_{\mathbf{h}_l}$ and the final answers executed $\mathcal{A}_{\mathbf{h}_l}=\text{Exec}(\text{Convert}(F_{\mathbf{h}_l}),\mathcal{G})$.

\begin{proposition}
The agent's awareness of the environment makes it more effective in generating optimal logical forms compared to end-to-end methods.
\end{proposition}
\begin{proof} 
We provide quantitative experimental results in Section~\ref{Section5.4} and qualitative proofs in Appendix~\ref{proof1}.
\end{proof}

\subsection{Heuristic Environment Exploration}
\label{Section3.2}
As shown in Figure~\ref{f3}, to address the issue of the step-by-step agent falling into local optima or large search spaces, we design an MCTS-based heuristic environment exploration method, driven by a policy model and a reward model.

\subsubsection{The Policy Model}
The policy model aims to provide the agent with a forward-looking capability.
We use the last state at each step $\mathbf{h}_{t-1}$ from the annotated training set $\mathcal{D}_a$ as input, and the steps from the current state to the conclusion $(\mathbf{e}_t,...,\mathbf{e}_l)$ as output, forming SFT data for training the policy model $\pi_{\text{policy}}$: 
\begin{equation}
\small
    \mathcal{L}_{\text{SFT}}(\pi_{\text{policy}}, \mathcal{D}_a) = - \mathbb{E}_{\mathcal{D}_a} \left[ \sum_{t=1}^{l} \log \pi_{\text{policy}} \left(\sum_{i=t}^{l}\mathbf{e}_i \mid \mathbf{h}_{t-1}\right) \right].
\end{equation}

\subsubsection{The Reward Model}
The reward model aims to assess the entire trajectory by scoring the final logical form.
We use the question $\mathcal{Q}$ as input, and the logical form $F_{\mathbf{h}_l}$ from the annotated training set $\mathcal{D}_a$ as output, forming SFT data for training the reward model $\pi_{\text{reward}}$:
\begin{equation}
\small
    \mathcal{L}_{\text{SFT}}(\pi_{\text{reward}}, \mathcal{D}_a) = - \mathbb{E}_{\mathcal{D}_a} \left[ \log \pi_{\text{reward}} \left(F_{\mathbf{h}_l} \mid \mathcal{Q}\right) \right],
\end{equation}

Moreover, we employ a scoring method $R_{\pi}$ that uses the logits from the LLM $\pi$, which can be $\pi_{\text{policy}}$ or $\pi_{\text{reward}}$, to evaluate the likelihood of an output $y$, given an input $x$:
\begin{equation}
\small
    R_{\pi}(y \mid x)=\beta +\alpha \log \pi\left(y \mid x\right).
\end{equation}
where $\beta$ is the defined full score, set as 100, and $\alpha$ is a positive temperature to control the disparity of scores.

\subsubsection{Monte Carlo Tree Search over KB}
MCTS is a heuristic search algorithm in the form of a tree, where each node represents a state in the agent process. Starting from the initial state (the root node), the algorithm uses four stages of selection, expansion, simulation, and back-propagation to explore and enrich the search tree iteratively. MCTS conducts a total of $N$ search rollouts. In the $n$-th rollout ($n=1,...,N$), the agent process can be represented by the trajectory of agent states $\{[\mathbf{h}^{(n)}_t]_{t=1}^l\}_{n=1}^N$.

\textbf{Selection. } When a new MCTS rollout begins, the agent process starts from the root node and progressively searches down through the child nodes of the already explored tree until it reaches a leaf node. At each level, the UCT (Upper Confidence Bound applied to Trees)~\citep{MCTS} algorithm is used to select the next child node:
\begin{equation}
\small
    \mathbf{e}_t \gets {\arg\max}_{\mathbf{e} \in E(\mathbf{h}^{(n)}_{t-1})} \left[ Q(\mathbf{h}^{(n)}_{t-1}+\mathbf{e}) + w \sqrt{\frac{\ln N(\mathbf{h}^{(n)}_{t-1})}{N(\mathbf{h}^{(n)}_{t-1}+\mathbf{e})}} \right],
\end{equation}
where $N(.)$ is the visit counts of the agent state during the MCTS process, $E(.)$ is the candidate expansion of Thought-Action-Observation explorations, derived from \Cref{E7}, and $Q(.)$ is the Q-value of the agent state, which will be updated by back-propagation. UCT balances the selection of high-scoring nodes with the exploration of unvisited ones. The variable $w$ controls the tendency towards exploration. A larger $w$ encourages exploration of nodes with fewer visits, while a smaller $w$ biases nodes with higher scores.

\textbf{Expansion. }
Once the selection process reaches a leaf node but not in a terminal \texttt{Finish} state and is not beyond the maximum depth $L$, the policy model $\pi_{\text{policy}}$ generates several possible next states by beam search:
\begin{equation}
\small
    \{\mathbf{e}_t^{(b)}\}_{b=1}^B\sim\pi_{\text{policy}}\left(\mathbf{e}_t \mid \mathbf{h}_{t-1}^{(n)}\right)_{\text{beam}},
\end{equation}
where $B$ is the beam size. To engage with KB environment, we utilize an unsupervised retrieval model, SimCSE~\citep{SimCSE}, to match the generated explorations with the exploration options $\mathbf{e} \in \text{Exp}(\mathbf{h}_{t-1}^{(n)},\mathcal{G})$ that are executable over KB $\mathcal{G}$ when connected with the last state $\mathbf{h}_{t-1}^{(n)}$:
\begin{equation}
\small
    \{\mathbf{e}_t^{(i)}\}_{i=1}^k\! \gets\! {\arg\max}^k_{\mathbf{e} \in \text{Exp}(\mathbf{h}_{t-1}^{(n)},\mathcal{G})}\text{SimCSE}\!\left(\{\mathbf{e}_t^{(b)}\}_{b=1}^B, \mathbf{e}\right).
\end{equation}
For example, if the ground truth is `\texttt{Find\_relation} [ file.actor.film ]' the model might initially generate `[ film.actor ]'. Then, we select the most semantically related action options from $\text{Exp}(\mathbf{h}_{t-1}^{(n)},\mathcal{G})$ such as `[ file.actor.film ]' and `[ tv.tv\_actor.starring\_roles ]' to filter the top $k$ explorations $\{\mathbf{e}_t^{(i)}\}_{i=1}^k$.

Then, the policy model $\pi_{\text{policy}}$ scores the $k$ candidates based on the previous state $\mathbf{h}_{t-1}^{(n)}$ and selects the top $d$ candidates as expanded options $E(\mathbf{h}_{t-1}^{(n)})$, which are added as child nodes to the leaf node, thus expanding the tree: 
\begin{equation}
\label{E7}
\small
    E(\mathbf{h}_{t-1}^{(n)})=\{\mathbf{e}_t^{(i)}\}_{i=1}^d \gets {\arg\max}^d R_{\pi_{\text{policy}}}\left(\{\mathbf{e}_t^{(i)}\}_{i=1}^k \mid \mathbf{h}_{t-1}^{(n)} \right).
\end{equation}

\textbf{Simulation. }
After the nodes are expanded, the policy model assigns scores to all newly added child nodes. The node with the highest prospective score is selected:
\begin{equation}
\small
    \mathbf{e}_{t} \gets {\arg\max}_{\mathbf{e} \in E(\mathbf{h}^{(n)}_{t-1})} R_{\pi_{\text{policy}}}\left(\mathbf{e} \mid \mathbf{h}^{(n)}_{t-1}\right),
\end{equation}
and the simulation continues to explore the process until the final \texttt{Finish} state, producing a complete logical form generation trajectory.

\textbf{Back-propagation. }
Once the final state is reached, the reward model $\pi_{\text{reward}}$ evaluates the entire trajectory by assessing the corresponding logical form combined with the policy model's score from the last step to compute the overall Q-value of the final state:
\begin{equation}
\small
    Q(\mathbf{h}^{(n)}_l) \gets \delta\ R_{\pi_{\text{policy}}}\left(\mathbf{e}_l \mid \mathbf{h}^{(n)}_{l-1}\right) + (1-\delta)\ R_{\pi_{\text{reward}}}\left(F_{\mathbf{h}^{(n)}_{l}} \mid \mathcal{Q}\right),
\end{equation}
where $\delta$ is a ratio from $(0,1)$ to balance the process score and overall score. 
The algorithm then back-propagates the score by updating the Q-values of all nodes along the trajectory, from the leaf back to the root:
\begin{equation}
\small
    Q(\mathbf{h}_t^{(n)}) \gets \max_{j=1}^n\left(\frac{\sum_{i=l}^{t} Q(\mathbf{h}_i^{(j)})}{l - t + 1}\right),
\end{equation}
where the Q-values of parent nodes are updated to the maximum average Q-value from all child nodes along the trajectory. Meanwhile, the visit count of each node along the trajectory $N(\mathbf{h}_t^{(n)})$ adds 1, and then the next rollout begins.

\subsubsection{Final Trajectory Chosen}
After MCTS completes its exploration for $N$ rollouts with parameter set $\theta$, we select the trajectory $\hat{\mathbf{h}}_l^{\mathcal{Q}}\in\{\mathbf{h}^{(n)}_l\}_{n=1}^N$ with the highest Q-value in every state in the expanded search tree as the optimal trajectory of question $\mathcal{Q}$:
\begin{equation}
\small
(\hat{\mathbf{h}}_l^{\mathcal{Q}},\hat{F}^{\mathcal{Q}},\hat{\mathcal{A}}^{\mathcal{Q}})=\text{MCTS}_{\theta}\left(\mathcal{Q},\pi_{\text{policy}},\pi_{\text{reward}}\right),
\end{equation}
where $\hat{F}^{\mathcal{Q}}$ is corresponding logical form of $\hat{\mathbf{h}}_l^{\mathcal{Q}}$, and $\hat{\mathcal{A}}^{\mathcal{Q}}=\text{Exec}(\text{Convert}(\hat{F}^{\mathcal{Q}}),\mathcal{G})$ is the executed answers.

\begin{proposition}
The MCTS-based heuristic method balances the effectiveness and size of the search space better than CoT-based and ToT-based step-by-step methods.
\end{proposition}
\begin{proof} 
We provide quantitative experimental results in Section~\ref{Section5.4} and qualitative proofs in Appendix~\ref{proof2}.
\end{proof}

\begin{table*}[t]
\caption{\label{t2}
40-shot results on the local dev set of GrailQA. \textbf{Bold} numbers indicate the best low-resource performance.
}
\fontsize{8pt}{8pt}\selectfont
\centering
\vskip 0.1in
\setlength{\tabcolsep}{2mm}{
\begin{tabular}{llcccccc|cc}
\toprule
\multirow{2.5}{*}{\textbf{Method}} & \multirow{2.5}{*}{\textbf{LLM}} & \multicolumn{2}{c}{\textbf{I.I.D}} & \multicolumn{2}{c}{\textbf{Compositional}} & \multicolumn{2}{c}{\textbf{Zero-shot}} & \multicolumn{2}{c}{\textbf{Overall}} \\ \cmidrule(lr){3-4} \cmidrule(lr){5-6} \cmidrule(lr){7-8} \cmidrule(lr){9-10}
 &  & \textbf{EM} & \textbf{F1} & \textbf{EM} & \textbf{F1} & \textbf{EM} & \textbf{F1} & \textbf{EM} & \textbf{F1} \\ \midrule
\multicolumn{10}{c}{\textit{Full-Supervised KBQA Methods}} \\ \midrule
RnG-KBQA~\cite{RnG-KBQA} &  & 86.7 & 89.0 & 61.7 & 68.9 & 68.8 & 74.7 & 69.5 & 76.9 \\ 
DecAF~\cite{DecAF} &  & 88.7 & 92.4 & 71.5 & 79.8 & 65.9 & 77.3 & 72.5 & 81.4 \\ 
TIARA~\cite{TIARA} &  & 88.4 & 91.2 & 66.4 & 74.8 & 73.3 & 80.7 & 75.3 & 81.9 \\ \midrule
\multicolumn{10}{c}{\textit{Low-resource KBQA Methods}} \\ \midrule
KB-BINDER~\cite{KB-BINDER} & GPT-3.5-turbo & 40.0 & 43.3 & 33.9 & 36.6 & 40.1 & 44.0 & 38.7 & 42.2 \\ 
KB-Coder~\cite{KB-Coder} & GPT-3.5-turbo & 40.6 & 45.5 & 34.5 & 38.6 & 42.2 & 47.3 & 40.1 & 44.9 \\ 
ARG-KBQA~\cite{ARG-KBQA} & GPT-3.5-turbo & 46.6 & 51.5 & 36.4 & 41.8 & 46.6 & 52.1 & 43.8 & 48.5 \\ \midrule
\multirow{2}{*}{KBQA-o1 (Llama-3)} & Llama-3.1-8B & 77.8 {\fontsize{5pt}{5pt}\selectfont ±0.5} & 85.5 {\fontsize{5pt}{5pt}\selectfont ±0.4} & 76.3 {\fontsize{5pt}{5pt}\selectfont ±0.6} & 77.6 {\fontsize{5pt}{5pt}\selectfont ±0.5} & 68.1 {\fontsize{5pt}{5pt}\selectfont ±0.8} & 76.1 {\fontsize{5pt}{5pt}\selectfont ±0.4} & 71.9 {\fontsize{5pt}{5pt}\selectfont ±0.3} & 78.5 {\fontsize{5pt}{5pt}\selectfont ±1.0} \\ 
 & Llama-3.3-70B & \textbf{79.2} {\fontsize{5pt}{5pt}\selectfont ±0.7} & \textbf{88.2} {\fontsize{5pt}{5pt}\selectfont ±0.3} & 80.8 {\fontsize{5pt}{5pt}\selectfont ±0.6} & 82.3 {\fontsize{5pt}{5pt}\selectfont ±0.2} & 71.8 {\fontsize{5pt}{5pt}\selectfont ±1.0} & 80.3 {\fontsize{5pt}{5pt}\selectfont ±0.8} & 75.2 {\fontsize{5pt}{5pt}\selectfont ±0.7} & 81.6 {\fontsize{5pt}{5pt}\selectfont ±0.5} \\ \midrule
\multirow{4}{*}{KBQA-o1 (Qwen2.5)} & Qwen2.5-7B & 76.1 {\fontsize{5pt}{5pt}\selectfont ±0.3} & 84.4 {\fontsize{5pt}{5pt}\selectfont ±0.5} & 75.7 {\fontsize{5pt}{5pt}\selectfont ±0.7} & 77.0 {\fontsize{5pt}{5pt}\selectfont ±0.7} & 67.3 {\fontsize{5pt}{5pt}\selectfont ±0.9} & 75.7 {\fontsize{5pt}{5pt}\selectfont ±0.2} & 70.8 {\fontsize{5pt}{5pt}\selectfont ±0.5} & 77.9 {\fontsize{5pt}{5pt}\selectfont ±0.8} \\ 
 & Qwen2.5-14B & 78.0 {\fontsize{5pt}{5pt}\selectfont ±0.4} & 85.9 {\fontsize{5pt}{5pt}\selectfont ±0.6} & 77.4 {\fontsize{5pt}{5pt}\selectfont ±0.8} & 78.4 {\fontsize{5pt}{5pt}\selectfont ±0.5} & 68.6 {\fontsize{5pt}{5pt}\selectfont ±0.1} & 79.0 {\fontsize{5pt}{5pt}\selectfont ±0.7} & 72.5 {\fontsize{5pt}{5pt}\selectfont ±0.6} & 79.2 {\fontsize{5pt}{5pt}\selectfont ±0.2} \\ 
 & Qwen2.5-32B & 78.9 {\fontsize{5pt}{5pt}\selectfont ±0.5} & 86.2 {\fontsize{5pt}{5pt}\selectfont ±0.7} & 79.3 {\fontsize{5pt}{5pt}\selectfont ±0.6} & 80.4 {\fontsize{5pt}{5pt}\selectfont ±0.8} & 69.6 {\fontsize{5pt}{5pt}\selectfont ±0.3} & 79.7 {\fontsize{5pt}{5pt}\selectfont ±0.7} & 73.3 {\fontsize{5pt}{5pt}\selectfont ±0.8} & 80.3 {\fontsize{5pt}{5pt}\selectfont ±1.3} \\ 
 & Qwen2.5-72B & 79.1 {\fontsize{5pt}{5pt}\selectfont ±0.4} & 87.4 {\fontsize{5pt}{5pt}\selectfont ±0.5} & \textbf{81.5} {\fontsize{5pt}{5pt}\selectfont ±0.7} & \textbf{83.0} {\fontsize{5pt}{5pt}\selectfont ±0.6} & \textbf{72.1} {\fontsize{5pt}{5pt}\selectfont ±0.4} & \textbf{81.9} {\fontsize{5pt}{5pt}\selectfont ±0.5} & \textbf{75.8} {\fontsize{5pt}{5pt}\selectfont ±0.4} & \textbf{82.1} {\fontsize{5pt}{5pt}\selectfont ±0.2} \\ \midrule
\multirow{2}{*}{KBQA-o1 (Gemma-2)} & Gemma-2-9B & 77.1 {\fontsize{5pt}{5pt}\selectfont ±0.5} & 82.3 {\fontsize{5pt}{5pt}\selectfont ±0.6} & 75.6 {\fontsize{5pt}{5pt}\selectfont ±0.8} & 76.7 {\fontsize{5pt}{5pt}\selectfont ±0.4} & 66.3 {\fontsize{5pt}{5pt}\selectfont ±0.2} & 76.4 {\fontsize{5pt}{5pt}\selectfont ±0.6} & 70.6 {\fontsize{5pt}{5pt}\selectfont ±0.7} & 77.8 {\fontsize{5pt}{5pt}\selectfont ±0.5} \\ 
 & Gemma-2-27B & 78.3 {\fontsize{5pt}{5pt}\selectfont ±0.3} & 82.6 {\fontsize{5pt}{5pt}\selectfont ±0.4} & 76.0 {\fontsize{5pt}{5pt}\selectfont ±0.6} & 77.3 {\fontsize{5pt}{5pt}\selectfont ±0.5} & 69.5 {\fontsize{5pt}{5pt}\selectfont ±1.1} & 79.6 {\fontsize{5pt}{5pt}\selectfont ±0.7} & 72.8 {\fontsize{5pt}{5pt}\selectfont ±0.5} & 79.7 {\fontsize{5pt}{5pt}\selectfont ±0.3} \\ \bottomrule
\end{tabular}}
\vspace{-3mm}
\end{table*}

\begin{table}[t]
\caption{\label{t3}
100-shot results on the test set of WebQSP. \textbf{Bold} numbers indicate the best low-resource performance.
}
\fontsize{8pt}{8pt}\selectfont
\centering
\vskip 0.1in
\setlength{\tabcolsep}{2mm}{
\begin{tabular}{llc}
\toprule
\textbf{Method} & \textbf{LLM} & \textbf{F1} \\ \midrule
\multicolumn{3}{c}{\textit{Full-Supervised KBQA Methods}} \\ \midrule
RnG-KBQA~\cite{RnG-KBQA} &  & 75.6 \\ 
DecAF~\cite{DecAF} &  & 76.7 \\ 
TIARA~\cite{TIARA} &  & 78.7 \\ \midrule
\multicolumn{3}{c}{\textit{Low-resource KBQA Methods}} \\ \midrule
KB-BINDER~\cite{KB-BINDER} & GPT-3.5-turbo & 52.6 \\ 
KB-Coder~\cite{KB-Coder} & GPT-3.5-turbo & 55.7 \\ 
ARG-KBQA~\cite{ARG-KBQA} & GPT-3.5-turbo & 58.8 \\ \midrule
\multirow{2}{*}{KBQA-o1 (Llama-3)} & Llama-3.1-8B & 59.8 {\fontsize{5pt}{5pt}\selectfont ±1.2} \\ 
 & Llama-3.3-70B & \textbf{67.0} {\fontsize{5pt}{5pt}\selectfont ±0.4} \\ \midrule
\multirow{4}{*}{KBQA-o1 (Qwen2.5)} & Qwen2.5-7B & 57.8 {\fontsize{5pt}{5pt}\selectfont ±0.7} \\ 
 & Qwen2.5-14B & 60.1 {\fontsize{5pt}{5pt}\selectfont ±1.3} \\ 
 & Qwen2.5-32B & 63.7 {\fontsize{5pt}{5pt}\selectfont ±0.9} \\ 
 & Qwen2.5-72B & 66.5 {\fontsize{5pt}{5pt}\selectfont ±1.1} \\ \midrule
\multirow{2}{*}{KBQA-o1 (Gemma-2)} & Gemma-2-9B & 58.9 {\fontsize{5pt}{5pt}\selectfont ±0.6} \\ 
 & Gemma-2-27B & 61.0 {\fontsize{5pt}{5pt}\selectfont ±0.5} \\ \bottomrule
\end{tabular}}
\vspace{-0.8mm}
\end{table}

\begin{table}[t]
\caption{\label{t4}
100-shot results on the test set of GraphQ. \textbf{Bold} numbers indicate the best low-resource performance.
}
\fontsize{8pt}{8pt}\selectfont
\centering
\vskip 0.1in
\setlength{\tabcolsep}{2mm}{
\begin{tabular}{llc}
\toprule
\textbf{Method} & \textbf{LLM} & \textbf{F1} \\ \midrule
\multicolumn{3}{c}{\textit{Full-Supervised KBQA Methods}} \\ \midrule
SPARQA~\cite{SPARQA} &  & 21.5 \\ 
BERT+Ranking~\cite{GrailQA} &  & 25.0 \\ 
ArcaneQA~\cite{ArcaneQA} &  & 31.8 \\ \midrule
\multicolumn{3}{c}{\textit{Low-resource KBQA Methods}} \\ \midrule
KB-BINDER~\cite{KB-BINDER} & GPT-3.5-turbo & 27.1 \\ 
KB-Coder~\cite{KB-Coder} & GPT-3.5-turbo & 31.1 \\ 
ARG-KBQA~\cite{ARG-KBQA} & GPT-3.5-turbo & - \\ \midrule
\multirow{2}{*}{KBQA-o1 (Llama-3)} & Llama-3.1-8B & 48.7 {\fontsize{5pt}{5pt}\selectfont ±0.8} \\ 
 & Llama-3.3-70B & 50.5 {\fontsize{5pt}{5pt}\selectfont ±0.2} \\ \midrule
\multirow{4}{*}{KBQA-o1 (Qwen2.5)} & Qwen2.5-7B & 49.2 {\fontsize{5pt}{5pt}\selectfont ±0.2} \\ 
 & Qwen2.5-14B & 50.0 {\fontsize{5pt}{5pt}\selectfont ±0.7} \\ 
 & Qwen2.5-32B & 50.9 {\fontsize{5pt}{5pt}\selectfont ±0.6} \\ 
 & Qwen2.5-72B & \textbf{51.2} {\fontsize{5pt}{5pt}\selectfont ±1.0} \\ \midrule
\multirow{2}{*}{KBQA-o1 (Gemma-2)} & Gemma-2-9B & 49.8 {\fontsize{5pt}{5pt}\selectfont ±0.7} \\ 
 & Gemma-2-27B & 50.3 {\fontsize{5pt}{5pt}\selectfont ±0.4} \\ \bottomrule
\end{tabular}}
\vspace{-0.8mm}
\end{table}

\subsection{Incremental Fine-Tuning}
\label{Section3.3}
In addition to training on annotated data, KBQA-o1 employs MCTS with exploration incentives $\theta_{\text{exp}}$ for heuristic exploration on unannotated questions $\mathcal{Q} \in \mathcal{D}_n$:
\begin{equation}
\small
    \{(\hat{\mathbf{h}}_l^{\mathcal{Q}},\hat{F}^{\mathcal{Q}},\hat{\mathcal{A}}^{\mathcal{Q}})\}_{\mathcal{Q} \in \mathcal{D}_n}\!=\!\left\{\text{MCTS}_{\theta_{\text{exp}}}\!\left(\mathcal{Q},\pi_{\text{policy}},\pi_{\text{reward}}\right)\right\}_{\mathcal{Q} \in \mathcal{D}_n}. 
\end{equation}
Then, we discard the annotation by choosing if the answer set is not empty and the reward score of logical form does not exceed a threshold $\gamma^*$:
\begin{equation}
\small
\hat{R}^{\mathcal{Q}}=R_{\pi_{\text{reward}}}\left(\hat{F}^{\mathcal{Q}} \mid \mathcal{Q}\right),
\end{equation}
\begin{equation}
\small
    \mathcal{D}_i=\mathcal{D}_a\cup\left\{\left(\mathcal{Q},\hat{F}^{\mathcal{Q}},\hat{\mathcal{A}}^{\mathcal{Q}}\right)|\hat{\mathcal{A}}^{\mathcal{Q}}\neq\emptyset\land \hat{R}^{\mathcal{Q}}>\gamma^*\right\}_{\mathcal{Q} \in \mathcal{D}_n}.
\end{equation}
Combined with the original annotated data $\mathcal{D}_a$, the incremental data $\mathcal{D}_i$ is then used for incremental fine-tuning of the policy and reward models:
\begin{equation}
\small
    \mathcal{L}_{\text{SFT}}(\pi_{\text{policy}}, \mathcal{D}_i) = - \mathbb{E}_{\mathcal{D}_i} \left[ \sum_{t=1}^{l} \log \pi_{\text{policy}} \left(\sum_{i=t}^{l}\mathbf{e}_i \mid \mathbf{h}_{t-1}\right) \right],
\end{equation}
\begin{equation}
\small
    \mathcal{L}_{\text{SFT}}(\pi_{\text{reward}}, \mathcal{D}_i) = - \mathbb{E}_{\mathcal{D}_i} \left[ \log \pi_{\text{reward}} \left(F_{\mathbf{h}_l} \mid \mathcal{Q}\right) \right].
\end{equation}
Through incremental fine-tuning, the policy and reward models gain enhanced understanding of the environment and a preference for high-reward logical form trajectories. Finally, we perform testing $\mathcal{D}_t$ under efficiency-focused MCTS parameter settings $\theta_{\text{eff}}$, yielding final answers $\hat{\mathcal{A}}^{\mathcal{Q}}$:
\begin{equation}
\small
    \{(\hat{\mathbf{h}}_l^{\mathcal{Q}},\hat{F}^{\mathcal{Q}},\hat{\mathcal{A}}^{\mathcal{Q}})\}_{\mathcal{Q} \in \mathcal{D}_t}\!=\!\left\{\text{MCTS}_{\theta_{\text{eff}}}\!\left(\mathcal{Q},\pi_{\text{policy}},\pi_{\text{reward}}\right)\right\}_{\mathcal{Q} \in \mathcal{D}_t}. 
\end{equation}

\begin{proposition}
There exists a reward threshold $\gamma^*<\beta$ such that incremental fine-tuning data, under the effect of the KB, can improve model performance.
\end{proposition}
\begin{proof} 
We provide quantitative experimental results in Section~\ref{Section5.5} and qualitative proofs in Appendix~\ref{proof3}.
\end{proof}

\section{Experiments}
This section presents the experimental setup, results, and analysis. We answer the following research questions (RQs):
\textbf{RQ1}: Does KBQA-o1 outperform other KBQA methods?
\textbf{RQ2}: Does the main component of KBQA-o1 work?
\textbf{RQ3}: Does KBQA-o1 address the corresponding challenges compared to end-to-end and step-by-step KBQA methods?
\textbf{RQ4}: How does incremental fine-tuning gradually improve low-resource KBQA performance?

\subsection{Experimental Setup}

\textbf{Datasets. } All experiments are conducted on three standard KBQA datasets in low-resource setting~\cite{KB-BINDER}: GrailQA~\cite{GrailQA}, WebQSP~\cite{WebQSP}, and GraphQ~\citep{GraphQ}. All datasets are based on Freebase~\citep{Freebase} KB. 

\textbf{Baselines. } We mainly compare our method with KB-BINDER~\cite{KB-BINDER}, KB-Coder~\cite{KB-Coder}, and ARG-KBQA~\cite{ARG-KBQA} with the same LLM of gpt-3.5-turbo-0613 from OpenAI in low-resource setting. Some results obtained by full-supervised training on the whole training dataset~\cite{RnG-KBQA,DecAF,TIARA,SPARQA,GrailQA,ArcaneQA} are also reported for reference.

\begin{table*}[t]
\caption{\label{t5}
Evaluation results of Llama-3.1-8B-based KBQA-o1 and its ablations. \textbf{Bold} numbers indicate the best performance.}
\fontsize{8pt}{8pt}\selectfont
\centering
\vskip 0.1in
\setlength{\tabcolsep}{2mm}{
\begin{tabular}{lcccccc|cc|c|c}
\toprule
\multirow{2.5}{*}{\textbf{Method}} & \multicolumn{2}{c}{\textbf{I.I.D}} & \multicolumn{2}{c}{\textbf{Compositional}} & \multicolumn{2}{c}{\textbf{Zero-shot}} & \multicolumn{2}{c}{\textbf{GrailQA}} & \multicolumn{1}{c}{\textbf{WebQSP}} & \multicolumn{1}{c}{\textbf{GraphQ}} \\ \cmidrule(lr){2-3} \cmidrule(lr){4-5} \cmidrule(lr){6-7} \cmidrule(lr){8-9} \cmidrule(lr){10-10} \cmidrule(lr){11-11} 
& \textbf{EM} & \textbf{F1} & \textbf{EM} & \textbf{F1} & \textbf{EM} & \textbf{F1} & \textbf{EM} & \textbf{F1} & \textbf{F1} & \textbf{F1} \\ \midrule
KBQA-o1 (Llama-3.1-8B)& \textbf{77.8} {\fontsize{5pt}{5pt}\selectfont ±0.5} & \textbf{85.5} {\fontsize{5pt}{5pt}\selectfont ±0.4} & \textbf{76.3} {\fontsize{5pt}{5pt}\selectfont ±0.6} & \textbf{77.6} {\fontsize{5pt}{5pt}\selectfont ±0.5} & \textbf{68.1} {\fontsize{5pt}{5pt}\selectfont ±0.8} & \textbf{76.1} {\fontsize{5pt}{5pt}\selectfont ±0.4} & \textbf{71.9} {\fontsize{5pt}{5pt}\selectfont ±0.3} & \textbf{78.5} {\fontsize{5pt}{5pt}\selectfont ±1.0} & \textbf{59.8} {\fontsize{5pt}{5pt}\selectfont ±1.2} & \textbf{48.7} {\fontsize{5pt}{5pt}\selectfont ±0.8} \\ \midrule
w/o \textit{agent prompt} & 65.2 {\fontsize{5pt}{5pt}\selectfont ±1.1} & 72.8 {\fontsize{5pt}{5pt}\selectfont ±0.8} & 61.9 {\fontsize{5pt}{5pt}\selectfont ±1.3} & 63.2 {\fontsize{5pt}{5pt}\selectfont ±1.2} & 45.7 {\fontsize{5pt}{5pt}\selectfont ±1.6} & 49.8 {\fontsize{5pt}{5pt}\selectfont ±1.4} & 51.2 {\fontsize{5pt}{5pt}\selectfont ±2.3} & 56.4 {\fontsize{5pt}{5pt}\selectfont ±3.1} & 52.3 {\fontsize{5pt}{5pt}\selectfont ±1.2} & 37.7 {\fontsize{5pt}{5pt}\selectfont ±2.2} \\ 
w/o \textit{KB environment} & 43.5 {\fontsize{5pt}{5pt}\selectfont ±1.4} & 51.3 {\fontsize{5pt}{5pt}\selectfont ±1.3} & 38.7 {\fontsize{5pt}{5pt}\selectfont ±1.7} & 41.5 {\fontsize{5pt}{5pt}\selectfont ±1.5} & 29.6 {\fontsize{5pt}{5pt}\selectfont ±1.9} & 35.4 {\fontsize{5pt}{5pt}\selectfont ±1.6} & 34.9 {\fontsize{5pt}{5pt}\selectfont ±2.0} & 43.0 {\fontsize{5pt}{5pt}\selectfont ±1.6} & 49.3 {\fontsize{5pt}{5pt}\selectfont ±0.7} & 25.1 {\fontsize{5pt}{5pt}\selectfont ±1.4} \\ \midrule
w/o \textit{initial annotated sft} & 14.3 {\fontsize{5pt}{5pt}\selectfont ±3.9} & 17.4 {\fontsize{5pt}{5pt}\selectfont ±4.2} & 12.2 {\fontsize{5pt}{5pt}\selectfont ±4.7} & 14.5 {\fontsize{5pt}{5pt}\selectfont ±4.1} & 8.5 {\fontsize{5pt}{5pt}\selectfont ±3.6} & 10.7 {\fontsize{5pt}{5pt}\selectfont ±4.4} & 10.6 {\fontsize{5pt}{5pt}\selectfont ±4.3} & 13.1 {\fontsize{5pt}{5pt}\selectfont ±4.0} & 17.2 {\fontsize{5pt}{5pt}\selectfont ±2.9} & 9.5 {\fontsize{5pt}{5pt}\selectfont ±3.0} \\
w/o \textit{MCTS} & 47.9 {\fontsize{5pt}{5pt}\selectfont ±1.6} & 54.2 {\fontsize{5pt}{5pt}\selectfont ±1.5} & 42.6 {\fontsize{5pt}{5pt}\selectfont ±2.1} & 45.8 {\fontsize{5pt}{5pt}\selectfont ±1.9} & 40.3 {\fontsize{5pt}{5pt}\selectfont ±1.8} & 44.2 {\fontsize{5pt}{5pt}\selectfont ±2.0} & 43.8 {\fontsize{5pt}{5pt}\selectfont ±2.6} & 48.5 {\fontsize{5pt}{5pt}\selectfont ±3.7} & 44.0 {\fontsize{5pt}{5pt}\selectfont ±2.1} & 20.8 {\fontsize{5pt}{5pt}\selectfont ±2.6}\\ \midrule
w/o \textit{incremental fine-tuning}& 60.7 {\fontsize{5pt}{5pt}\selectfont ±1.0} & 68.3 {\fontsize{5pt}{5pt}\selectfont ±1.2} & 56.2 {\fontsize{5pt}{5pt}\selectfont ±1.5} & 58.9 {\fontsize{5pt}{5pt}\selectfont ±1.4} & 53.8 {\fontsize{5pt}{5pt}\selectfont ±1.7} & 56.7 {\fontsize{5pt}{5pt}\selectfont ±1.5} & 57.1 {\fontsize{5pt}{5pt}\selectfont ±1.0} & 63.9 {\fontsize{5pt}{5pt}\selectfont ±1.6} & 55.4 {\fontsize{5pt}{5pt}\selectfont ±2.4} & 43.9 {\fontsize{5pt}{5pt}\selectfont ±1.7} \\ \bottomrule
\end{tabular}}
\vspace{-0.4mm}
\end{table*}

\begin{figure*}[t]
\footnotesize
\centering
\subfigure[\label{f4a}]{
    \includegraphics[width=0.33\textwidth]{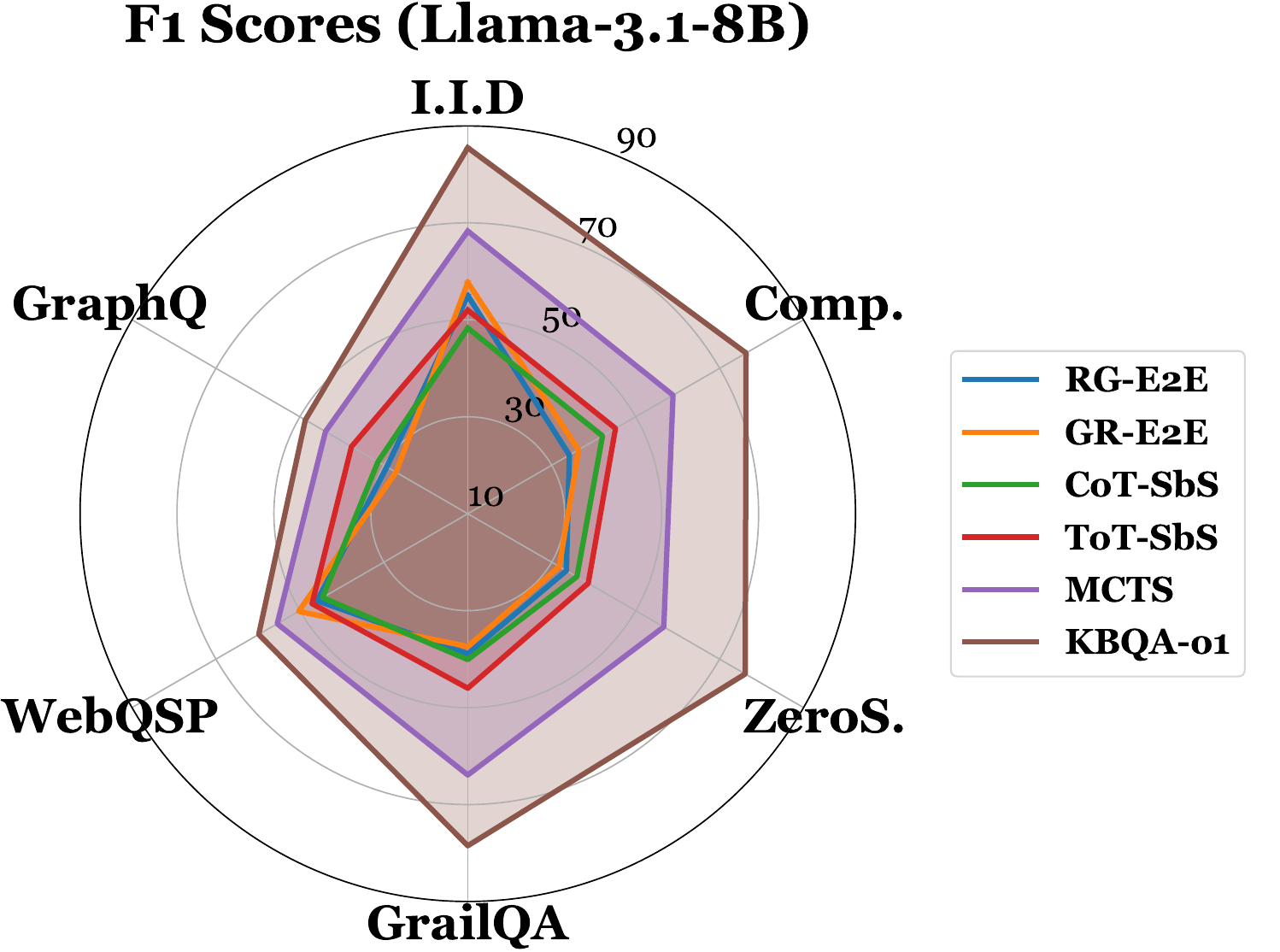}
}
\subfigure[\label{f4b}]{
    \includegraphics[width=0.33\textwidth]{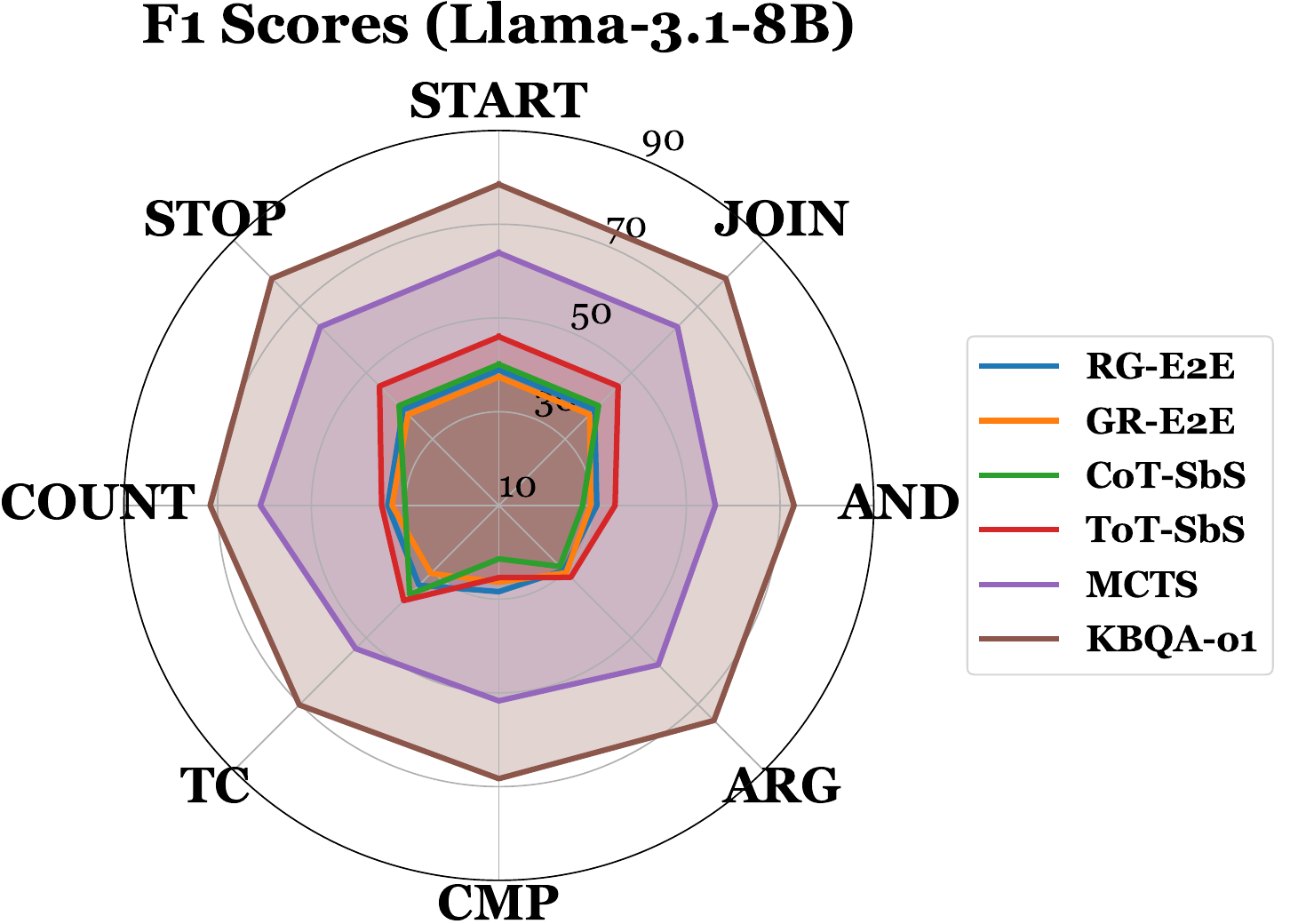}
}
\subfigure[\label{f4c}]{
    \includegraphics[width=0.29\textwidth]{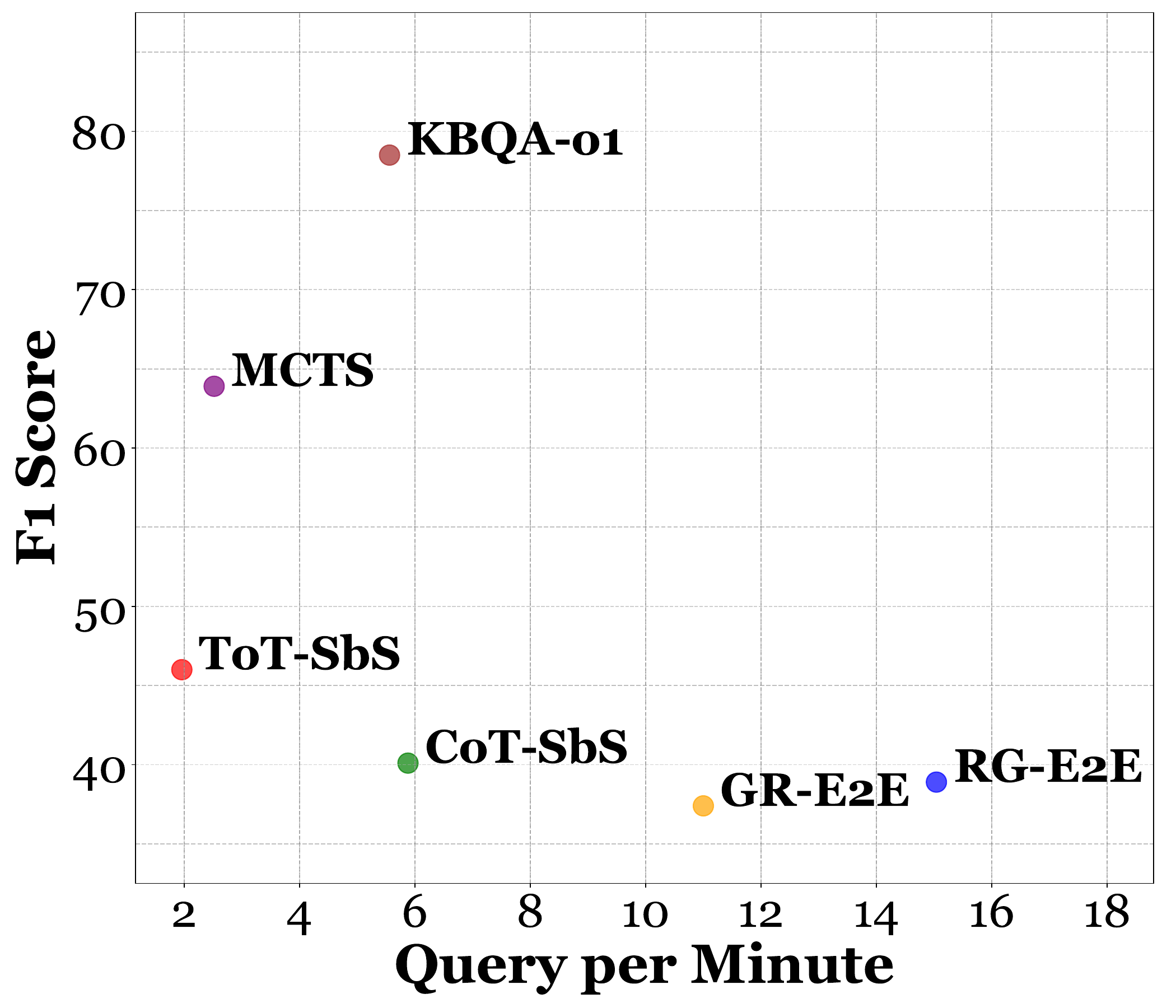}
}
\vspace{-1.7mm}
\caption{Performance and efficiency comparison of Llama-3.1-8B-based KBQA-o1 with compared methods. (a) F1 scores comparison across datasets. (b) F1 scores across logical operators on GrailQA. (c) Trade-off between F1 scores and queries per minute on GrailQA.}
\end{figure*}

\begin{figure*}[t]
\footnotesize
\centering
\subfigure[\label{f5a}]{
    \includegraphics[width=0.3\textwidth]{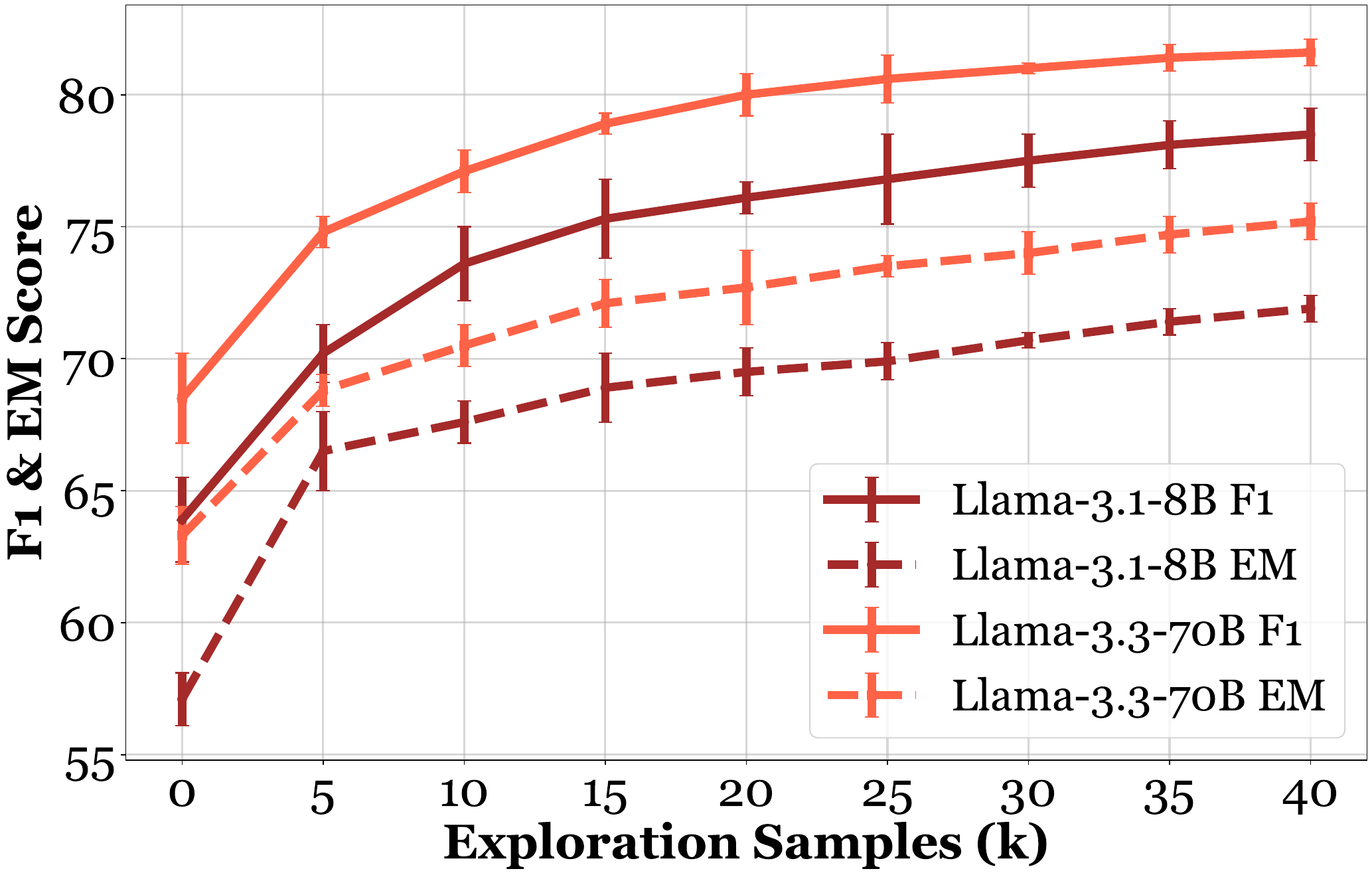}
}
\subfigure[\label{f5b}]{
    \includegraphics[width=0.3\textwidth]{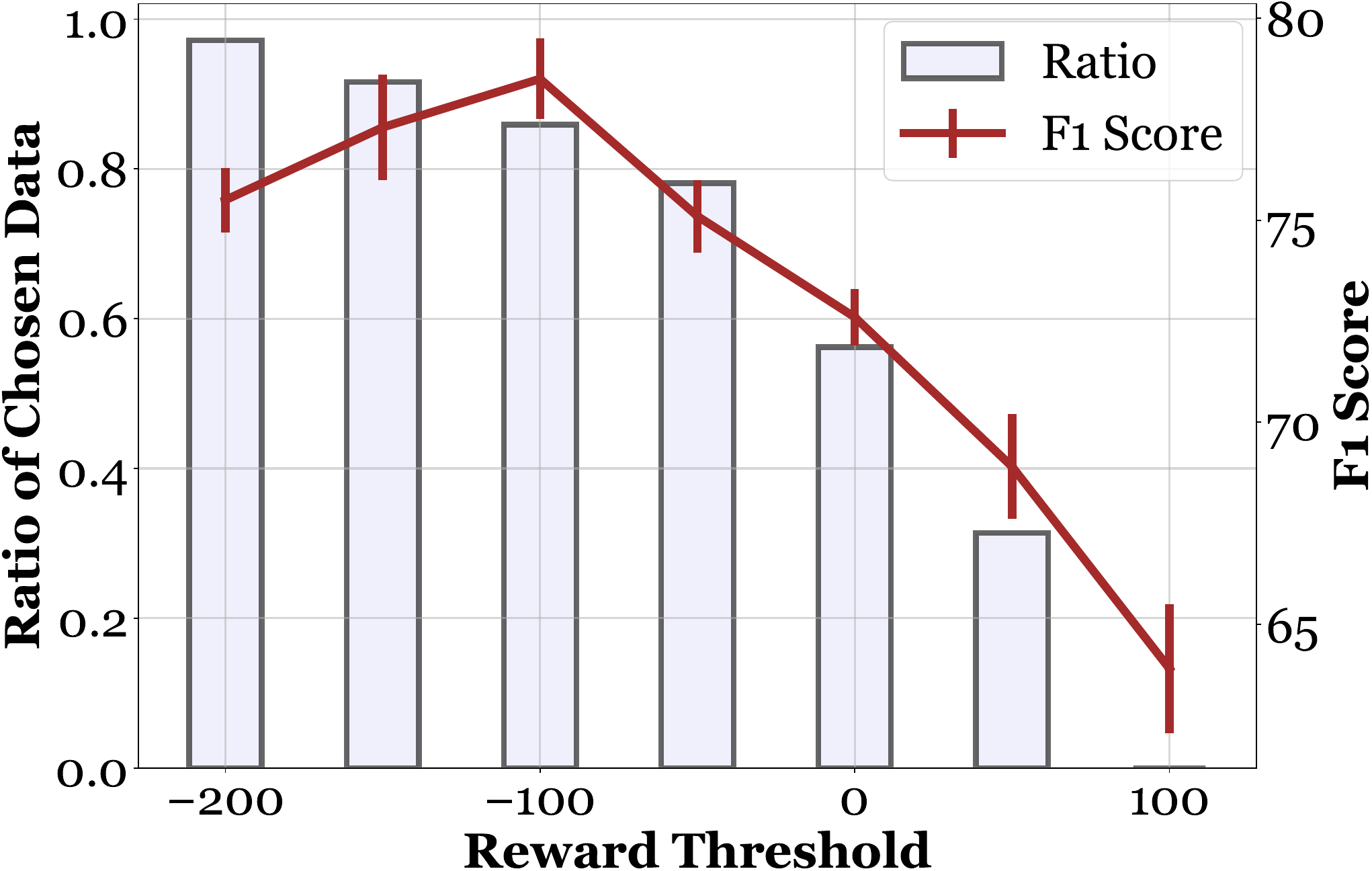}
}
\subfigure[\label{f5c}]{
    \includegraphics[width=0.35\textwidth]{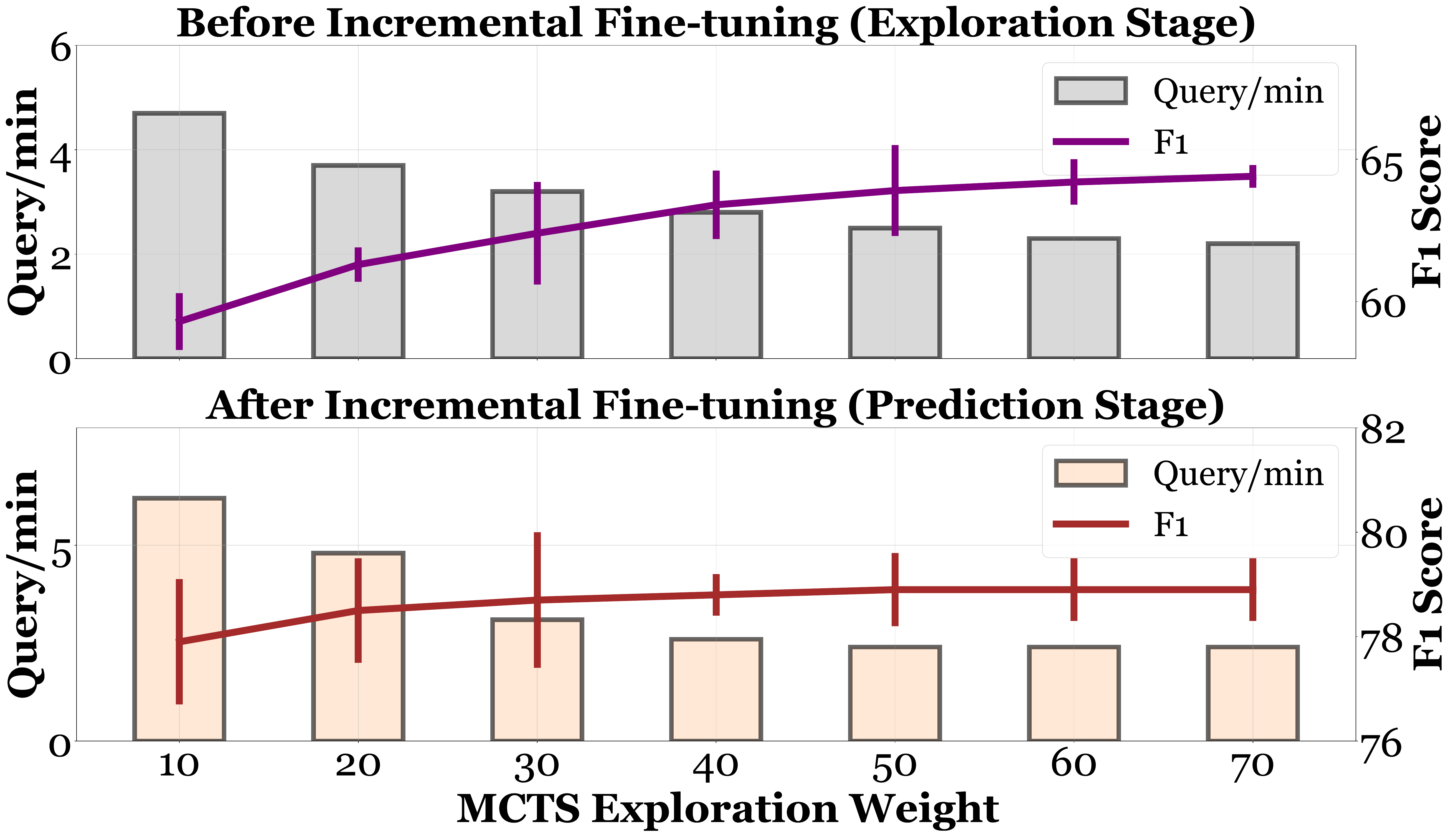}
}
\vspace{-1.7mm}
\caption{Impact of incremental fine-tuning tested on GrailQA: (a) Effect of exploration samples on F1 and EM scores. (b) Relationship between reward threshold, data ratio, and performance. (c) Influence of MCTS exploration weight on query efficiency and accuracy.}
\end{figure*}

\textbf{Evaluation Metrics. }
In line with prior studies~\citep{KB-BINDER,KB-Coder,ARG-KBQA}, we use F1 score and Exact Match (EM) as the evaluation metric for GrailQA, while F1 score are reported for WebQSP and GraphQ.

\textbf{Implementation Details. }
Following KB-BINDER~\citep{KB-BINDER}, we conduct 40-shot experiments for GrailQA, and 100-shot for WebQSP and GraphQ. 
During the MCTS exploration phase, we set $\theta_{\text{exp}}$ with $w=50$, while in the prediction phase, we set  $\theta_{\text{eff}}$ with $w = 10$. 
We select multiple open-source 7B-72B LLMs, including Llama-3~\cite{llama3}, Qwen2.5~\cite{Qwen2.5}, and Gemma-2~\cite{Gemma2}, to construct KBQA-o1.
All experiments are done on 8 NVIDIA A40 GPUs (48GB), with results averaged from three randomly seeded experiments. 
Appendix~\ref{hyper} shows the optimal hyperparameter settings.

\subsection{Main Result (RQ1)}
As shown in Tables \ref{t2}, \ref{t3}, and \ref{t4}, KBQA-o1 enables open-source LLM to outperform previous low-resource KBQA methods based on GPT-3.5-turbo, with limited labeled data. In more complex data sets such as GrailQA, KBQA-o1 improves the overall EM performance of the Llama-3.1-8B model by 28.1 percentage points and boosts F1 scores by 30.0 percentage points over the previous best methods. In compositional and zero-shot evaluations, KBQA-o1 even outperforms fully supervised KBQA methods, which demonstrates its strong capabilities for generalization, environment exploration, and handling complex logical questions. Moreover, KBQA-o1 is plug-and-play, allowing integration with various open-source 7B-72B LLMs, expected to improve further with future open-source LLM updates.

\subsection{Ablation Study (RQ2)}
As shown in Table~\ref{t5}, we conduct an ablation study on the Llama-3.1-8B-based KBQA-o1 methods. Five ablation settings are tested: removing the ReAct-based agent prompt, removing environment feedback, excluding the initial supervised fine-tuning (SFT) with a small amount of labeled data, removing MCTS optimization, and omitting the incremental fine-tuning stage, respectively. Comparisons of the evaluation results reveal that all modules contribute to overall performance, underscoring the importance of designed agent initialization, heuristic environment exploration, and incremental fine-tuning as key components of KBQA-o1.

\subsection{Comparison Analysis (RQ3)}
\label{Section5.4}
To explore how KBQA-o1 addresses the challenges of end-to-end and step-by-step methods mentioned in Section~\ref{Section1}, we construct six variants.
We design two end-to-end variants: one is a retrieve-then-generate method (RG-E2E) based on DECAF~\cite{DecAF}, and the other is a generate-then-retrieve method (GR-E2E), based on ChatKBQA~\cite{ChatKBQA}. Furthermore, we design two step-by-step variants: one is based on CoT (CoT-SbS) as QueryAgent~\cite{QueryAgent}, and the other is based on ToT (ToT-SbS) as ToG~\cite{ToG}. Finally, we implement an MCTS-optimized variant without incremental fine-tuning and the full KBQA-o1 method after incremental fine-tuning.

\textbf{Comparison with End-to-end Methods. }
To investigate whether the agent process in KBQA-o1 is more capable of perceiving the KB environment than end-to-end methods, we compare CoT-SbS with RG-E2E and GR-E2E. Under the same prompt, CoT-SbS is equivalent to the agent process in KBQA-o1 without expansion.
As shown in Figures \ref{f4a} and \ref{f4b}, in the I.I.D and WebQSP evaluations with annotations in distribution, RG-E2E and GR-E2E perform better. However, in Compositional and Zero-shot evaluations, the agent process represented by CoT-SbS performs better.
This is because, with preannotated logical forms, end-to-end generation rigidly adheres to a fixed logical form structure. In contrast, the step-by-step agent approach allows stepwise adjustments by KB environment awareness, enabling it to handle more complex out-of-distribution questions.

\textbf{Comparison with Step-by-step Methods. }
To investigate whether the MCTS proposed in KBQA-o1 can further prevent the step-by-step agent process from getting stuck in local optima or dealing with a large search space, we compared MCTS with CoT-SbS and ToT-SbS.
As shown in Figures \ref{f4a} and \ref{f4b}, MCTS consistently outperform CoT-SbS and ToT-SbS in all evaluation tasks and logical operators.
To analyze efficiency, Figure \ref{f4c} shows that the efficiency of MCTS and KBQA-o1 falls between CoT-SbS and ToT-SbS. When exploring unlabeled data, the MCTS prioritizes correctness, while the full KBQA-o1, after incremental fine-tuning, adopts more efficient settings to balance performance and search time.

\subsection{Analysis of Incremental Improvement (RQ4)}
\label{Section5.5}
To explore the impact of incremental fine-tuning on improving the performance of KBQA-o1 in low-resource KBQA scenarios, we examine three key factors: the impacts of the exploration samples, the reward threshold, and the exploration weights on performance and efficiency.

\textbf{Impact of Exploration Samples. }
As shown in Figure~\ref{f5a}, we gradually increase the number of unlabeled exploration samples and observe a steady improvement in the F1 and EM scores after incremental fine-tuning. This suggests that newly explored and labeled samples contribute to automatically generating high-quality logical forms using both environmental and reward-based filtering, allowing KBQA to improve with minimal human annotation, making it highly promising for transferability and real-world applications.

\textbf{Impact of Reward Threshold $\gamma^*$. }
As shown in Figure~\ref{f5b}, we conduct experiments with reward thresholds $\gamma^*$ ranging from -200 to 100. As $\gamma^*$ increases, the proportion of post-labeled data decreases until no data are selected at the maximum threshold of 100. In GrailQA, setting $\gamma^*$ to -100 retained most labeled samples while filtering out lower-quality ones, resulting in the highest F1 score. This indicates that selecting an appropriate $\gamma^*$ helps preserve newly labeled high-quality samples while discarding erroneous samples, making incremental fine-tuning effective.

\textbf{Impact of Exploration Weight $w$. }
Figure~\ref{f5c} presents a study on the selection of the exploration weight $w$ in MCTS before and after incremental fine-tuning. We test seven different values between 10 and 70. The results show that higher $w$ leads to slower generation, but improves accuracy. Based on this, we set $w\in\theta_{\text{exp}}$ at 50 before incremental fine-tuning to ensure accuracy. After incremental fine-tuning, we reduce $w\in\theta_{\text{eff}}$ to 10 to balance efficiency and effectiveness.
Hence, we ensure that the MCTS with $\theta_{\text{exp}}$ performs a slow and careful search during exploration, leading to high-quality data for incremental fine-tuning. After that, we can implement a more efficient MCTS with $\theta_{\text{eff}}$ in applications, maintaining high performance while improving efficiency.

\section{Conclusion}

In this work, we propose KBQA-o1, an agentic KBQA method with Monte Carlo Tree Search (MCTS) for efficient exploration. By combining a ReAct-based agent process with incremental fine-tuning, it improves logical form generation and reduces reliance on annotated data. Experiments on three KBQA datasets show that KBQA-o1 outperforms previous low-resource methods and rivals fully supervised models, demonstrating its scalability and effectiveness.

\section*{Acknowledgments}
This work is supported by the National Natural Science Foundation of China (Grant No. 62176026, Grant No. 62473271, and Grant No. 62406036). This work is also supported by the BUPT Excellent Ph.D. Students Foundation (No. CX2023133) and the Engineering Research Center of Information Networks, Ministry of Education, China.

\section*{Impact Statement}
This work introduces KBQA-o1, a novel agentic approach to knowledge base question answering (KBQA) using Monte Carlo Tree Search (MCTS). While effective, it still faces limitations in fine-grained policy design and scalability to larger domains. To address these challenges, we outline four future directions: (1) exploring reinforcement learning for continual learning, (2) enhancing logical reasoning capabilities, (3) adapting to specialized domains such as medicine and law, and (4) extending to multimodal, multilingual, and multi-agent settings, as detailed in Appendix~\ref{future}. This work poses no ethical concerns, as it relies solely on publicly available datasets and aims to advance KBQA technology with positive societal impacts.

\nocite{langley00}

\bibliography{example_paper}
\bibliographystyle{icml2025}

\newpage
\appendix
\onecolumn
\section*{Appendix} 

\section{Prompts Used in KBQA-o1}
\subsection{Initial Prompt}
\label{prompt1}

\begin{figure}[h!t]
\centering
\includegraphics[width=0.91\linewidth]{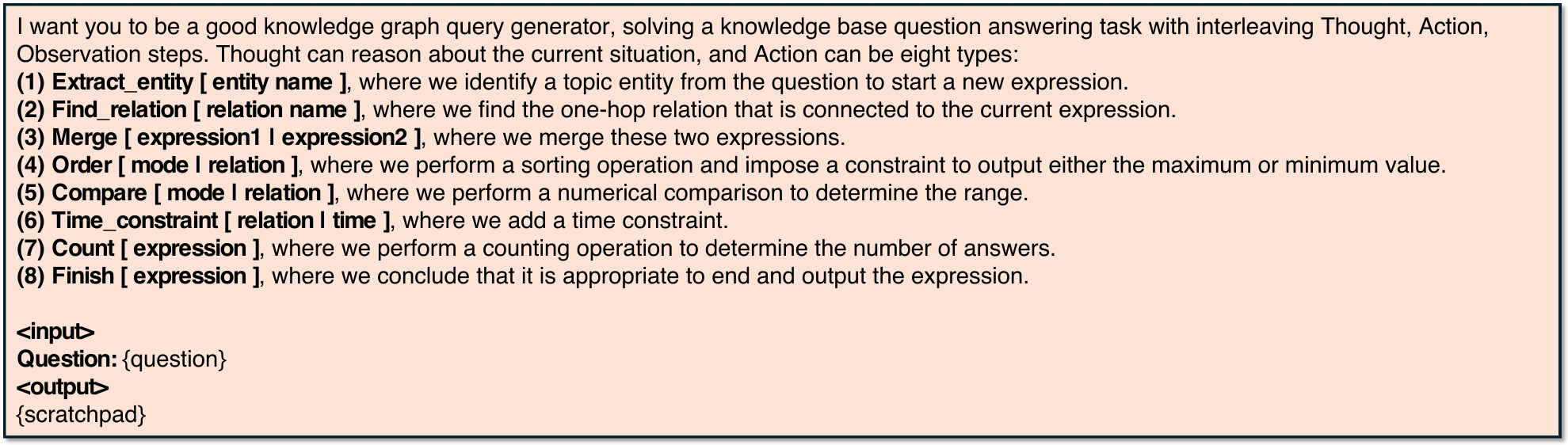}
\vspace{-1mm}
\caption{\label{fp1}
Initial prompt of agent process.}
\vspace{-4mm}
\end{figure}

Figure~\ref{fp1} illustrates the initial prompt used in the KBQA-o1 agent process. This prompt defines the structure and steps for solving KBQA tasks through interleaving Thought, Action, and Observation stages. The prompt guides the agent to perform specific actions, such as entity extraction, relation finding, expression merging, ordering, numerical comparisons, and adding time constraints. These predefined actions are essential for generating logical forms step by step from natural language questions. The diagram highlights the structured input format (\textless{}input\textgreater{} for the question and \textless{}output\textgreater{} for logical expressions) and the scratchpad used for intermediate reasoning steps during the task.

\subsection{Example Agent Process}
\label{prompt2}

\begin{figure}[h!t]
\centering
\includegraphics[width=0.91\linewidth]{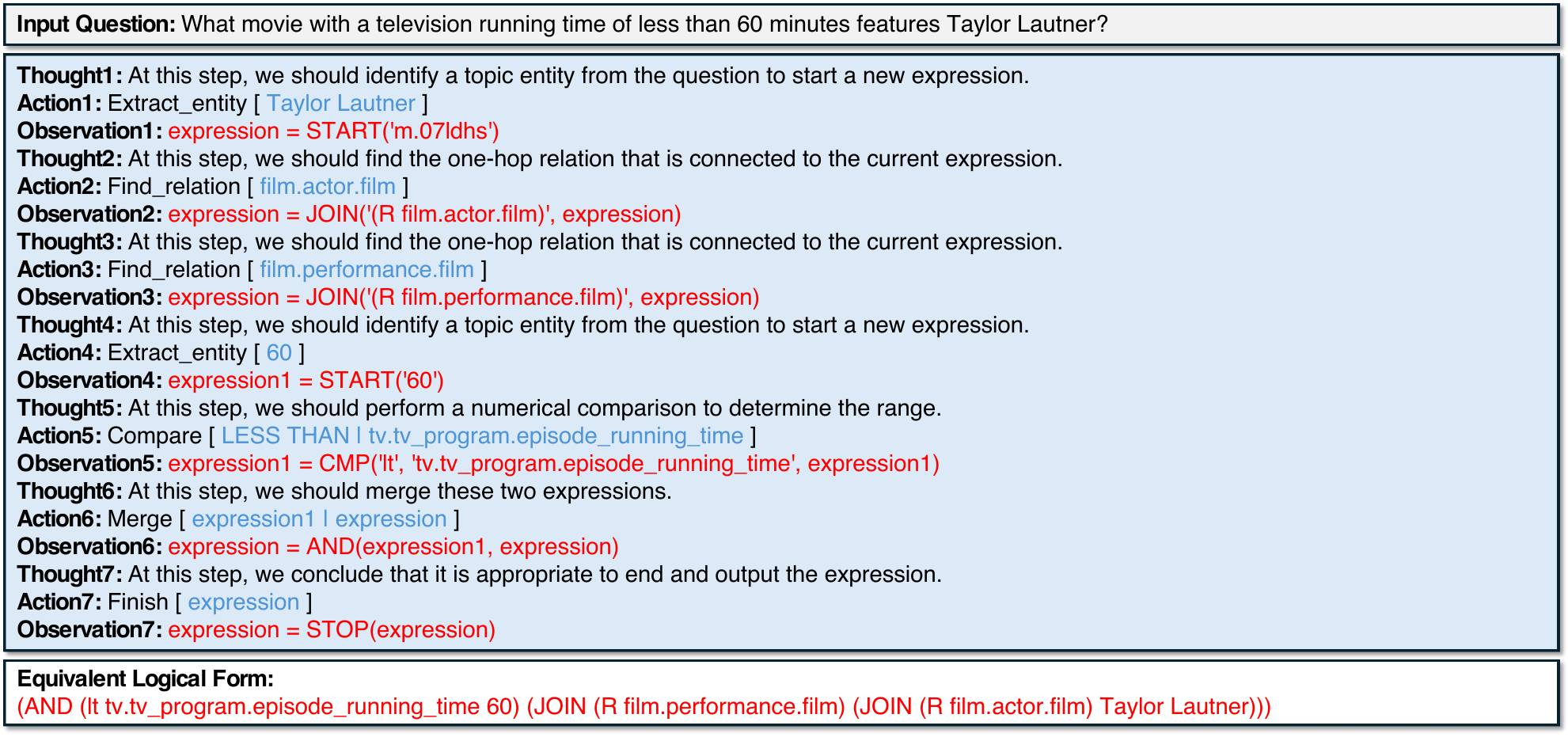}
\vspace{-1mm}
\caption{\label{fp2}
An example of a complete agent process.}
\vspace{-1mm}
\end{figure}

Figure~\ref{fp2} provides an example of the complete agent process in KBQA-o1 for the query. The figure demonstrates how the agent iteratively constructs the logical form using the Thought-Action-Observation prompt. Each step involves reasoning about the current context (Thought), performing a specific operation (Action), and observing the resulting logical expression (Observation). The process begins with extracting the topic entity followed by finding relevant relations and applying numerical and time constraints. The agent merges expressions to form a complete logical form, which is then executed to retrieve the answer from the knowledge base. The final logical output form is shown at the bottom, showcasing the agent’s ability to generate structured and executable queries systematically.

\section{Proof}
\subsection{Proof of Proposition 4.1}
\label{proof1}
\textbf{Proposition 4.1.}
\textit{The agent's awareness of the environment makes it more effective in generating optimal logical forms compared to end-to-end methods.}
\begin{proof} 
According to the Lyapunov Stability Second Theorem~\cite{Lyapunov}: 

For a dynamic system $\dot{x} = f(x)$, where $x$ is the state and $f(x)$ is the nonlinear function describing state evolution. Define a non-negative scalar function $V(x)$  such that: $V(x) > 0$ for all $x \neq 0$ and $V(0) = 0$ (positive definiteness). Its time derivative along the system trajectory satisfies $\dot{V}(x)<0$ and is bounded (negative definiteness). Then $V(x)$ is called a Lyapunov function candidate, and the system (from Lyapunov’s perspective) is asymptotically stable.

For a discrete dynamic system  $x_{t+1} = f(x_t)$, if a Lyapunov function $V(x)$ can be define, and the Lyapunov increment $\Delta V(x_t) = V(x_{t+1}) - V(x_t)$, then $V(x_t) > 0$  and $\Delta V(x_t) < 0$ with bounded implies asymptotic stability.

Our KBQA-o1 system is modeled as a discrete dynamic system:
\begin{equation}
h_{t+1} = f(h_t, \mathcal{G}),
\end{equation}
where $h_t \in \mathcal{H}$ is the system state at step $t$; $\mathcal{G}$ is environment feedback.
The state transition probability is defined as: $P(h_{t+1} \mid h_t, \mathcal{G})$, which represents the probability of transitioning to the next state $h_{t+1}$, given the current state $h_t$ and the environment feedback $\mathcal{G}$.
The goal is for the system state $h_t$ to converge asymptotically to the target state $h^*$, to satisfy $P(h^* \mid h_t, \mathcal{G}) = 1$.

We define a positive definite Lyapunov function $V(h_t)$ as:
\begin{equation}
V(h_t) = -\log P(h_t \mid h^*, \mathcal{G}),
\end{equation}
where $P(h_t \mid h^*, \mathcal{G})$ is the posterior probability of state $h_t$ relative to the target state $h^*$; $V(h_t)$ quantifies the deviation of the current state $h_t$ from the target state $h^*$.

The posterior probability satisfies $0 < P(h_t \mid h^*, \mathcal{G}) \leq 1$. Thus, $V(h_t) = -\log P(h_t \mid h^*, \mathcal{G}) \geq 0$.
When $h_t = h^*$, $P(h_t \mid h^*, \mathcal{G}) = 1 \quad \implies \quad V(h_t) = 0$.
Thus, $V(h_t)$ is positive definite.

The change in the Lyapunov function between steps is defined as:
\begin{equation}
\Delta V(h_t) = V(h_{t+1}) - V(h_t).
\end{equation}
Using the definition of $V(h_t)$, we get $V(h_{t+1})$:
\begin{equation}
V(h_{t+1}) = -\log P(h_{t+1} \mid h^*, \mathcal{G}),
\end{equation}
and the posterior probability recursion:
\begin{equation}
P(h_{t+1} \mid h^*, \mathcal{G}) = P(h_t \mid h^*, \mathcal{G}) \cdot P(h_{t+1} \mid h_t, \mathcal{G}),
\end{equation}
we have:
\begin{equation}
V(h_{t+1}) = -\log P(h_t \mid h^*, \mathcal{G}) - \log P(h_{t+1} \mid h_t, \mathcal{G}).
\end{equation}
Thus:
\begin{equation}
\Delta V(h_t) = V(h_{t+1}) - V(h_t) = -\log P(h_{t+1} \mid h_t, \mathcal{G}).
\end{equation}
The probability $P(h_{t+1} \mid h_t, \mathcal{G})$ can be expressed as:
\begin{equation}
P(h_{t+1} \mid h_t, \mathcal{G}) = P(\text{Exp}(h_t, \mathcal{G}) \mid h_t, \mathcal{G}) \cdot P(h_{t+1} \mid h_t, \text{Exp}(h_t, \mathcal{G})),
\end{equation}
where $P(\text{Exp}(h_t, \mathcal{G}) \mid h_t, \mathcal{G})$ is the probability that the environment generates an exploration space $\text{Exp}(h_t, \mathcal{G})$, containing valid actions for transitioning to $h_{t+1}$. And $P(h_{t+1} \mid h_t, \text{Exp}(h_t, \mathcal{G}))$ is the probability that the system transitions from $h_t$ to $h_{t+1}$, given the exploration space.

In KBQA-o1, our environment can provide all possible explorations under the prior state, and the correct exploration step is guaranteed to be included. Thus, $P(\text{Exp}(h_t, \mathcal{G}) \mid h_t, \mathcal{G}) = 1$. Therefore, $P(h_{t+1} \mid h_t, \mathcal{G}) = P(h_{t+1} \mid h_t, \text{Exp}(h_t, \mathcal{G}))$, and given a set of explorations that contains the correct one, the probability of selecting the correct exploration satisfies $0 < P(h_{t+1} \mid h_t, \text{Exp}(h_t, \mathcal{G})) < 1$. Consequently, $0 < P(h_{t+1} \mid h_t, \mathcal{G}) < 1$.

Since $0 < P(h_{t+1} \mid h_t, \mathcal{G}) < 1$, we conclude $\Delta V(h_t) < 0$ and is bounded. Thus, the Lyapunov function change is strictly negative definite.

Based on the Lyapunov Stability Second Theorem, $V(h_t) = -\log P(h_t \mid h^*, \mathcal{G})$ satisfies positive definiteness: $V(h_t) \geq 0$, and $V(h_t) = 0$ only when $h_t = h^*$, and negative definiteness: $\Delta V(h_t) < 0$, ensuring $V(h_t)$ strictly decreases. Thus, the Agent system is asymptotically stable in the Lyapunov sense, and the state $h_t$ converges to $h^*$.

However, for end-to-end methods, although both pre-retrieval and post-retrieval approaches can incorporate environmental information, pre-retrieval may result in an exploration space that does not include the target exploration, leading to $P(\text{Exp}(h_t, \mathcal{G}) \mid h_t, \mathcal{G}) = 0$. Post-retrieval, on the other hand, may generate a logical form framework that is incorrect, resulting in $P(h_{t+1} \mid h_t, \text{Exp}(h_t, \mathcal{G})) = 0$. In such cases, this leads to $P(h_{t+1} \mid h_t, \mathcal{G}) = P(\text{Exp}(h_t, \mathcal{G}) \mid h_t, \mathcal{G}) \cdot P(h_{t+1} \mid h_t, \text{Exp}(h_t, \mathcal{G})) = 0$, causing $\Delta V(h_t)$ isn't bounded, leading to instability. 

Therefore, we conclude that the agent’s awareness of the environment makes it more effective in generating optimal logical forms compared to end-to-end methods.

\end{proof}
\subsection{Proof of Proposition 4.2}
\label{proof2}
\textbf{Proposition 4.2.}
\textit{The MCTS-based heuristic method balances the effectiveness and size of the search space better than CoT-based and ToT-based step-by-step methods.}
\begin{proof} 
Let $\mathcal{F}$ represent the search space of logical forms, with a size of $\lvert \mathcal{F} \rvert = k^L$, where $k$ is the number of choices per step, and $L$ is the maximum search depth. Let $\mathcal{F}_{\text{optimal}} \subseteq \mathcal{F}$ denote the set of high-quality solutions whose scores exceed a threshold $\tau$. For any search method $X$, if its generated candidate set is $\mathcal{F}_X \subseteq \mathcal{F}$, the coverage rate is defined as:
\begin{equation}
C_X = \frac{\lvert \mathcal{F}_X \cap \mathcal{F}_{\text{optimal}} \rvert}{\lvert \mathcal{F}_{\text{optimal}} \rvert}.
\end{equation}

We first compare CoT (Chain-of-Thought) and our proposed MCTS-based method. CoT follows a single path $\mathbf{h}_l$, producing a single candidate $F_{\text{CoT}}$, with its candidate set given by $\mathcal{F}_{\text{CoT}} = \{ F_{\text{CoT}} \}$. If $F_{\text{CoT}} \in \mathcal{F}_{\text{optimal}}$, the coverage rate is
\begin{equation}
C_{\text{CoT}} = \frac{1}{\lvert \mathcal{F}_{\text{optimal}} \rvert},
\end{equation}
otherwise, $C_{\text{CoT}} = 0$.
MCTS, on the other hand, expands multiple different paths during multiple rollouts. If $N$ rollouts are performed in total, the resulting candidate set is
\begin{equation}
\mathcal{F}_{\text{MCTS}} = \bigcup_{n=1}^N \{ \mathbf{h}_l^{(n)} \},
\end{equation}
where $\mathbf{h}_l^{(n)}$ is the n-th path generated. Under effective strategy and reward guidance, most paths will fall into $\mathcal{F}_{\text{optimal}}$, and thus
\begin{equation}
C_{\text{MCTS}} = \frac{\lvert \mathcal{F}_{\text{MCTS}} \cap \mathcal{F}_{\text{optimal}} \rvert}{\lvert \mathcal{F}_{\text{optimal}} \rvert} \gg C_{\text{CoT}}.
\end{equation}
This demonstrates that MCTS significantly outperforms CoT in terms of coverage of high-quality solutions.

Next, we compare MCTS and ToT (Tree of Thought) based on BFS/DFS. ToT explores all possible paths in $\mathcal{F}$, so its max candidate set is $\mathcal{F}_{\text{ToT}} = \mathcal{F}$. The coverage rate for ToT is
\begin{equation}
C_{\text{ToT}} = \frac{\lvert \mathcal{F} \cap \mathcal{F}_{\text{optimal}} \rvert}{\lvert \mathcal{F}_{\text{optimal}} \rvert} = 1,
\end{equation}
but this requires a time complexity of
\begin{equation}
T_{\text{ToT}} = O(k^L).
\end{equation}
MCTS, however, only explores part of the tree. Assuming each rollout expands $\omega k$ candidates ($\omega \in (0, 1)$), and N rollouts are performed, the total search size is $N \cdot \omega k \cdot L$, with a time complexity of
\begin{equation}
T_{\text{MCTS}} = O(N \cdot \omega k \cdot L).
\end{equation}
As long as $N \cdot \omega k \cdot L \ll k^L$, we have
\begin{equation}
T_{\text{MCTS}} \ll T_{\text{ToT}}.
\end{equation}
Moreover, if the strategy and reward models effectively focus the search on high-potential nodes,
\begin{equation}
\lvert \mathcal{F}_{\text{MCTS}} \cap \mathcal{F}_{\text{optimal}} \rvert \approx \lvert \mathcal{F}_{\text{optimal}} \rvert,
\end{equation}
leading to
\begin{equation}
C_{\text{MCTS}} \approx C_{\text{ToT}} = 1.
\end{equation}

Combining these results, we see that compared to CoT, MCTS significantly improves the coverage rate by expanding multiple paths. Compared to ToT, MCTS achieves a near-equal coverage rate with substantially lower time complexity. Thus, MCTS balances effectiveness (coverage rate) and efficiency (search space size) better than both CoT and ToT.
\end{proof}

\subsection{Proof of Proposition 4.3}
\label{proof3}
\textbf{Proposition 4.3.}
\textit{There exists a reward threshold $\gamma^*<\beta$ such that incremental fine-tuning data, under the joint effect of the KB and the reward model, can significantly improve model performance.}
\begin{proof} 
First, we define the performance improvement of incremental fine-tuning, $\Delta E(\gamma^*)$. The model’s initial performance on the annotated dataset $\mathcal{D}_a$ is denoted as $E_{\text{base}}$, and its performance after incremental fine-tuning is $E_{\text{new}} = E_{\text{base}} + \Delta E(\gamma^*). Here, \Delta E(\gamma^*)$ represents the performance gain contributed by incremental data, which depends on the quantity and quality of the incremental data.

The incremental data originates from the unannotated dataset $\mathcal{D}_n$, where the logical forms $\hat{F}^{\mathcal{Q}}$ and answers $\hat{\mathcal{A}}^{\mathcal{Q}}$ are generated through a filtering mechanism. The filtering criteria are as follows: 1)	The reward score of the logical form $R_{\pi_{\text{reward}}}(F) > \gamma^*$; 2) The answer set is non-empty, $\hat{\mathcal{A}}^{\mathcal{Q}} \neq \emptyset$.

The logical forms that satisfy the criteria form the incremental dataset $\mathcal{D}_i(\gamma^*)$. The size and quality of the incremental data are both dependent on the reward threshold $\gamma^*$.
The performance improvement can be expressed as:
\begin{equation}
\Delta E(\gamma^*) = \lambda \cdot N_{\text{high}}(\gamma^*) \cdot Q_{\text{avg}}(\mathcal{D}_i(\gamma^*)),
\end{equation}
where $N_{\text{high}}(\gamma^*)$ is the quantity of incremental data; $Q_{\text{avg}}(\mathcal{D}_i(\gamma^*))$ is the average quality of incremental data; $\lambda > 0$ is a scaling factor representing the contribution of data quality to the performance gain.

Next, we analyze the properties of $N_{\text{high}}(\gamma^*)$ and $Q_{\text{avg}}(\mathcal{D}_i(\gamma^*))$, and subsequently derive the behavior of $\Delta E(\gamma^*)$.

The quantity of incremental data $N_{\text{high}}(\gamma^*)$ depends on the number of logical forms in the unannotated dataset that satisfy $R_{\pi_{\text{reward}}}(F) > \gamma^*$. Let $N_n = |\mathcal{D}n|$ be the size of the unannotated dataset, and let the reward scores of logical forms follow a probability density function $P(R)$. Then:
\begin{equation}
N_{\text{high}}(\gamma^*) = N_n \cdot P_{\text{high}}(\gamma^*),
\end{equation}
where $P_{\text{high}}(\gamma^*)$ represents the probability that the reward score exceeds $\gamma^*$:
\begin{equation}
P_{\text{high}}(\gamma^*) = \int_{\gamma^*}^{\beta} P(R) \, dR.
\end{equation}
Clearly, as $\gamma^*$ increases, $N_{\text{high}}(\gamma^*)$ decreases. When $\gamma^* \to -\infty$, $N_{\text{high}}(\gamma^*) \to N_n; when \gamma^* \to \beta, N_{\text{high}}(\gamma^*) \to 0$.

The quality of the incremental data $Q_{\text{avg}}(\mathcal{D}i(\gamma^*))$ is the average reward score of logical forms with $R{\pi_{\text{reward}}}(F) > \gamma^*$, defined as:
\begin{equation}
Q_{\text{avg}}(\mathcal{D}_i(\gamma^*)) = \frac{\int_{\gamma^*}^{\beta} R \cdot P(R) \, dR}{\int_{\gamma^*}^{\beta} P(R) \, dR}.
\end{equation}
The numerator $\int_{\gamma^*}^{\beta} R \cdot P(R) \, dR$ is the total quality weighted by reward scores, while the denominator $\int_{\gamma^*}^{\beta} P(R) \, dR = P_{\text{high}}(\gamma^*)$ is the total probability.
When $\gamma^* \to -\infty$, all logical forms are included in the incremental data, and the average quality is the expected reward score:
\begin{equation}
Q_{\text{avg}}(\mathcal{D}_i(\gamma^*)) \to \mathbb{E}[R] = \int_{-\infty}^{\beta} R \cdot P(R) \, dR.
\end{equation}
When $\gamma^* \to \beta$, only the logical forms with the highest reward scores are retained, and the average quality approaches 1.

Now, we analyze the behavior of the performance improvement $\Delta E(\gamma^*)$. Substituting $N_{\text{high}}(\gamma^*)$ and $Q_{\text{avg}}(\mathcal{D}i(\gamma^*))$ into $\Delta E(\gamma^*)$, we obtain:
\begin{equation}
\Delta E(\gamma^*) = \lambda \cdot N_n \cdot \left(\int_{\gamma^*}^{\beta} P(R) \, dR\right) \cdot \frac{\int_{\gamma^*}^{\beta} R \cdot P(R) \, dR}{\int_{\gamma^*}^{\beta} P(R) \, dR}.
\end{equation}
Simplifying this, we get:
\begin{equation}
\Delta E(\gamma^*) = \lambda \cdot N_n \cdot \int_{\gamma^*}^{\beta} R \cdot P(R) \, dR.
\end{equation}
To analyze the critical points, we take the derivative of $\Delta E(\gamma^*)$ with respect to $\gamma^*$:
\begin{equation}
\frac{\partial \Delta E(\gamma^*)}{\partial \gamma^*} = -\lambda \cdot N_n \cdot \gamma^* \cdot P(\gamma^*).
\end{equation}
It is clear that when $\gamma^* \to -\infty$, $\frac{\partial \Delta E(\gamma^*)}{\partial \gamma^*} > 0$, indicating that $\Delta E(\gamma^*)$ increases with $\gamma^*$; when $\gamma^* \to \beta$, $\frac{\partial \Delta E(\gamma^*)}{\partial \gamma^*} < 0$, indicating that $\Delta E(\gamma^*)$ decreases with $\gamma^*$.
Thus, $\Delta E(\gamma^*)$ is a unimodal function, and there exists a critical point $\gamma^*_{\text{opt}} \in (-\infty, \beta)$ at which $\Delta E(\gamma^*)$ achieves its maximum.

In conclusion, we have proven that: There exists a reward threshold $\gamma^*<\beta$ such that the trade-off between the quantity and quality of incremental data is optimized. 
At this threshold, the performance improvement $\Delta E(\gamma^*)$ reaches its maximum.
This establishes the existence of $\gamma^*$ and its role in optimizing the effect of incremental fine-tuning.

\end{proof}

\section{MCTS Algorithm Details}
\label{implement}

\begin{figure}[h!t]
\centering
\includegraphics[width=0.9\linewidth]{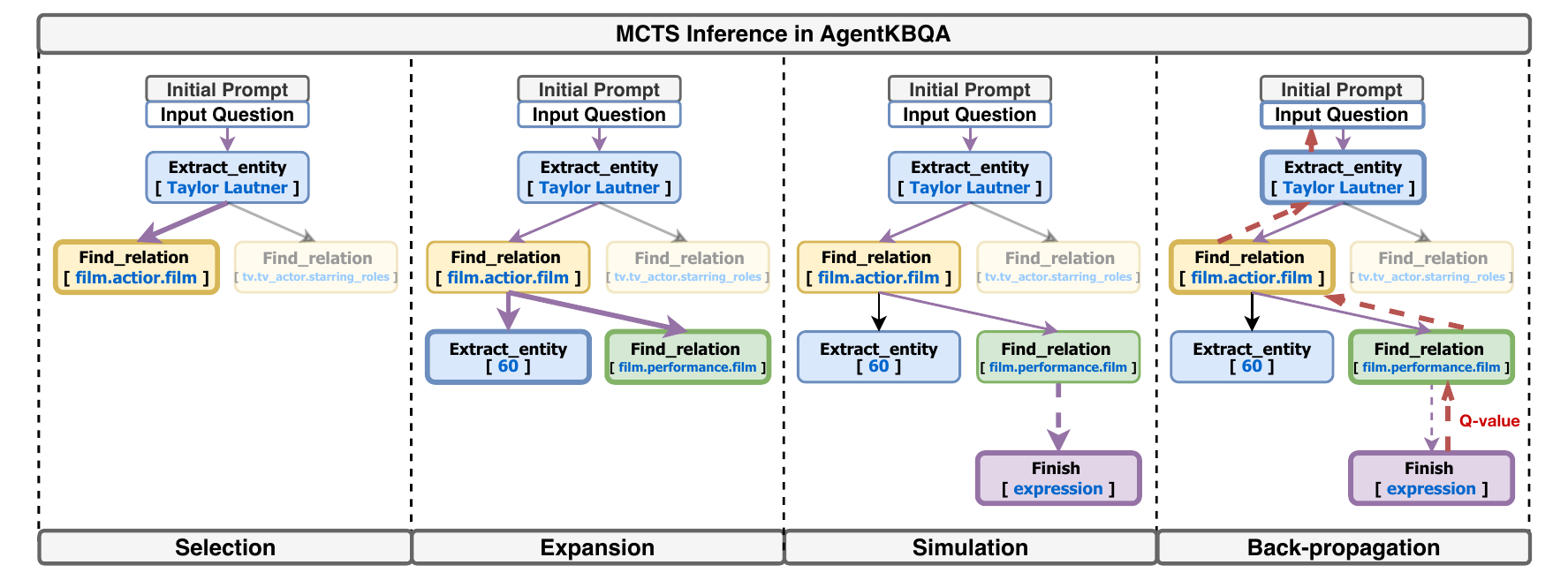}
\caption{Four stages of the MCTS process in KBQA-o1.}
\label{f_mcts}
\end{figure}

Figure~\ref{f_mcts} and \Cref{alg_mcts} illustrate the Monte Carlo Tree Search (MCTS) process in KBQA-o1. The figure highlights the four stages of MCTS: Selection, where nodes are chosen using the Upper Confidence Bound for Trees (UCT) to balance exploration and exploitation; Expansion, where candidate actions are generated by the policy model, filtered for relevance to the knowledge base, and added as child nodes; Simulation, where the most promising path is explored to produce a complete logical form and compute rewards; and Back-propagation, where rewards are propagated back to update Q-values and visit counts. The pseudocode formalizes this process, iteratively performing rollouts that follow the four stages. It ensures efficient exploration by selecting nodes with high potential, expanding with semantically relevant actions, simulating logical forms, and updating scores through back-propagation. This approach enables KBQA-o1 to navigate large search spaces effectively and generate high-quality logical forms.

\textbf{Complexity Analysis. }
The time complexity of the MCTS process in KBQA-o1 can be analyzed based on its four stages: Selection, Expansion, Simulation, and Back-propagation. In the Selection stage, the algorithm traverses the search tree up to depth  $L$ , selecting the best node using the Upper Confidence Bound for Trees (UCT), leading to a complexity of  $O(k \cdot L)$  per rollout, where  k  is the number of possible actions per step. The Expansion stage generates  B  beam candidates, which are filtered based on knowledge base similarity, adding  $O(\omega k)$  complexity. The Simulation stage explores paths up to depth  L , contributing  $O(k \cdot L)$ . Finally, the Back-propagation stage updates the rewards along the path, requiring  $O(L)$ . Summing up these steps, the complexity for a single rollout is  $O(k \cdot L)$ , and with  $N$  rollouts, the total complexity of MCTS is  $O(N \cdot k \cdot L)$ , which scales linearly with the number of rollouts and search depth.

Compared to exhaustive search methods like Tree-of-Thoughts (ToT), which have a complexity of  $O(k^L)$ , MCTS is significantly more efficient. Instead of exploring all possible paths, MCTS selectively explores high-potential nodes based on the policy model and UCT, reducing redundant computations. This selective approach allows MCTS to scale effectively, making it suitable for large KBs. Additionally, MCTS dynamically balances exploration and exploitation, adapting its search strategy based on reward feedback. This adaptability, combined with its lower computational cost, enables MCTS to generate high-quality logical forms while maintaining efficiency.

\begin{algorithm}[h!t]
\small
\caption{MCTS in KBQA-o1}
\label{alg_mcts}
\begin{algorithmic}[1]
    \REQUIRE Initial state $\mathbf{h}_0$, knowledge base $\mathcal{G}$, policy model $\pi_{\text{policy}}$, reward model $\pi_{\text{reward}}$, beam size $B$, number of top candidates $d$, depth limit $L$, number of rollouts $N$, exploration weight $w$.
    \FOR{$n \gets 1$ to $N$}
        \STATE $t \gets 0$
        \WHILE{$N(\mathbf{h}_t) > 0$}
            \STATE Select by UCT: $\mathbf{e}_t \gets {\arg\max}_{\mathbf{e} \in E(\mathbf{h}^{(n)}_{t-1})} \left[ Q(\mathbf{h}^{(n)}_{t-1}+\mathbf{e}) + w \sqrt{\frac{\ln N(\mathbf{h}^{(n)}_{t-1})}{N(\mathbf{h}^{(n)}_{t-1}+\mathbf{e})}} \right]$  \hfill $\triangleright$ Selection
            \STATE Update $\mathbf{h}_{t}^{(n)} \gets \mathbf{h}_{t-1}^{(n)} + \mathbf{e}_t$
            \STATE $t \gets t + 1$
        \ENDWHILE
        \WHILE{$\mathbf{h}_t$ is not a terminal state $\land\ t \leq L$}
            \STATE Beam search by policy model: $\{\mathbf{e}_t^{(b)}\}_{b=1}^B\sim\pi_{\text{policy}}\left(\mathbf{e}_t \mid \mathbf{h}_{t-1}^{(n)}\right)_{\text{beam}}$  \hfill $\triangleright$ Expansion
            \STATE Semantic selection from KB: $\{\mathbf{e}_t^{(i)}\}_{i=1}^k\! \gets\! {\arg\max}^k_{\mathbf{e} \in \text{Exp}(\mathbf{h}_{t-1}^{(n)},\mathcal{G})}\text{SimCSE}\!\left(\{\mathbf{e}_t^{(b)}\}_{b=1}^B, \mathbf{e}\right)$
            \STATE Policy model rerank: $E(\mathbf{h}_{t-1}^{(n)})=\{\mathbf{e}_t^{(i)}\}_{i=1}^d \gets {\arg\max}^d R_{\pi_{\text{policy}}}\left(\{\mathbf{e}_t^{(i)}\}_{i=1}^k \mid \mathbf{h}_{t-1}^{(n)} \right)$
            \STATE Simulate along the best: $\mathbf{e}_{t} \gets {\arg\max}_{\mathbf{e} \in E(\mathbf{h}^{(n)}_{t-1})} R_{\pi_{\text{policy}}}\left(\mathbf{e} \mid \mathbf{h}^{(n)}_{t-1}\right)$   \hfill $\triangleright$ Simulation
            \STATE Update $\mathbf{h}_{t}^{(n)} \gets \mathbf{h}_{t-1}^{(n)} + \mathbf{e}_t$
            \STATE $t \gets t + 1$
        \ENDWHILE
        \STATE $l \gets t$
        \STATE Calculate reward: $Q(\mathbf{h}^{(n)}_l) \gets \delta\ R_{\pi_{\text{policy}}}\left(\mathbf{e}_l \mid \mathbf{h}^{(n)}_{l-1}\right) + (1-\delta)\ R_{\pi_{\text{reward}}}\left(F_{\mathbf{h}^{(n)}_{l}} \mid \mathcal{Q}\right)$
        \FOR{$t \gets l, \dots, 0$}
            \STATE Update $Q(\mathbf{h}_t^{(n)}) \gets \max_{j=1}^n\left(\frac{\sum_{i=l}^{t} Q(\mathbf{h}_i^{(j)})}{l - t + 1}\right)$ \hfill $\triangleright$ Back-propagation
            \STATE Add visit count: $N(\mathbf{h}_t^{(n)}) \gets N(\mathbf{h}_t^{(n)}) + 1$ 
        \ENDFOR
    \ENDFOR
\end{algorithmic}
\end{algorithm}

\section{Dataset Details}
\label{dataset}

Table~\ref{t_data} provides an overview of the dataset statistics used in the KBQA-o1 experiments across different settings: I.I.D (Independent and Identically Distributed), Compositional, Zero-shot, and three datasets (GrailQA, WebQSP, and GraphQ). The table includes the following rows: \textbf{\#Train} is the number of training examples used for each dataset. \textbf{\#Exploration} is the number of exploration queries generated during the heuristic exploration phase for each dataset. The numbers indicate the extensive exploration. \textbf{\#Test} is the number of testing examples for evaluation in each setting.

\begin{table}[h!t]
\caption{\label{t_data}
Dataset statistics of KBQA-o1.}
\vskip 0.1in
\centering
\setlength{\tabcolsep}{2mm}{
\begin{tabular}{lcccccc}
\toprule
               & \textbf{I.I.D} & \textbf{Compositional} & \textbf{Zero-shot} & \textbf{GrailQA} & \textbf{WebQSP} & \textbf{GraphQ} \\ \midrule
\textbf{\#Train }       & -     & -             & -         & 40      & 100     & 100     \\
\textbf{\#Exploration } & -     & -             & -         & 43851   & 2929   & 2332   \\
\textbf{\#Test}         & 1564  & 1487          & 3645      & 6696    & 1566   & 2319   \\ \bottomrule
\end{tabular}}
\end{table}

\textbf{GrailQA dataset}~\citep{GrailQA} is a large-scale dataset specifically designed to evaluate KBQA models across three key generalization levels: i.i.d., compositional, and zero-shot. It contains 64,331 questions, of which 6,696 in the local dev set with three levels are used for testing in the KBQA-o1 experiments. This dataset requires models to understand a wide range of logical forms and multi-hop reasoning over knowledge bases. Its exploration phase involves a significant number of queries (43,851), emphasizing the importance of comprehensive environment exploration. The dataset is particularly challenging in its compositional and zero-shot settings, where models need to handle unseen combinations of entities and relations, testing their ability to generalize beyond simple patterns.

\textbf{WebQSP dataset}~\citep{WebQSP} is an enriched version of the original WebQuestions dataset, providing semantic parses for each of its 4,737 questions. In KBQA-o1 experiments, 1,566 questions are used for testing, and 2,929 queries are generated during the exploration phase. This dataset focuses on evaluating the semantic understanding of KBQA models, requiring them to map natural language questions to precise logical forms. The inclusion of semantic parses enhances the dataset’s utility for training and evaluating models, enabling fine-grained analysis of their ability to disambiguate and retrieve information from the knowledge base. The relatively smaller size of WebQSP compared to GrailQA makes it a valuable benchmark for assessing model efficiency in low-resource settings.

\textbf{GraphQ dataset}~\citep{GraphQ} is a dataset aimed at testing KBQA models on complex reasoning tasks involving graph-structured data. It contains over 5,000 questions, with 2,319 used for testing in the KBQA-o1 experiments. The exploration phase generates 2,332 queries, reflecting the intricate nature of reasoning required by the dataset. GraphQ challenges models to navigate multi-hop relationships and resolve complex dependencies between entities, pushing them to understand the underlying structure of the knowledge graph. Its focus on characteristic-rich queries ensures a thorough evaluation of a model’s reasoning capabilities, making it an essential dataset for advancing KBQA methods.

\section{Atomic Query Tool Details}
As shown in Table~\ref{t1}, KBQA-o1 introduces eight atomic query tools designed to facilitate logical form generation for KBQA. These tools are tailored to systematically convert natural language queries into logical forms by leveraging the structural properties of KB. Each tool serves a specific function, ensuring precise interactions with the KB to retrieve or manipulate information. Below is an explanation of each tool:

\textbf{Extract\_entity} is used to identify and extract a specific entity from the knowledge base. It takes an entity as an argument and initializes the logical form with this entity. The target function is represented as START(`entity'), and the corresponding logical form is simply the entity itself. For instance, extracting ``Taylor Lautner” initializes the logical form with the entity identifier for the actor.

\textbf{Find\_relation} identifies a relation connected to the current logical form. It takes a relation as an argument and appends it to the existing logical form using the JOIN operation. The resulting logical form is (JOIN relation (expression)). For example, finding the film.actor.film relation connects an actor to their associated films.

\textbf{Merge} combines two logical expressions into a single conjunctive expression. It takes two arguments, expression1 and expression, and returns AND(expression1, expression) as the target function. The equivalent logical form is (AND (expression1) (expression)). For example, merging two conditions like ``actor is Taylor Lautner” and ``runtime $<$ 60 minutes” forms a combined condition.

\textbf{Order} determines the extreme value of a property, such as the maximum or minimum. It takes a mode (e.g., max or min) and a relation as arguments. The target function is ARG(`mode', expression, `relation'), and the logical form is (mode (expression) relation). For example, finding the longest film associated with an actor uses this tool.

\textbf{Compare} evaluates a property against a specified value using a comparison operator (e.g., $<$, $<=$, $>$, $>=$). It takes a mode and a relation as arguments. The target function is CMP(`mode', `relation', expression), and the corresponding logical form is (mode relation (expression)). For instance, checking if a film’s runtime is less than 60 minutes employs this tool.

\textbf{Time\_constraint} imposes a time-based constraint on the logical form. It takes a relation and a time as arguments, applying a temporal filter to the expression. The target function is TC(expression, `relation', `time'), and the logical form is (TC (expression) relation time). For example, filtering films released before a specific year uses this tool.

\textbf{Count} calculates the number of entities that satisfy a given condition. It takes an expression as an argument, returning the target function COUNT(expression) and the logical form (COUNT (expression)). For instance, counting the number of films an actor has appeared in utilizes this tool.

\textbf{Finish} signals the termination of the logical form generation process. It takes an expression as its argument and finalizes the logical form as STOP(expression). The corresponding logical form is simply (expression). This tool ensures the logical form is complete and ready for execution against the KB.

\section{Baseline Details}
\label{baseline}
The six full-resource baselines provide a comprehensive evaluation framework for KBQA. These methods leverage fully annotated datasets and advanced reasoning mechanisms to achieve high performance on various KBQA tasks. Below is a summary of each baseline:

\textbf{RnG-KBQA}~\cite{RnG-KBQA} is a retrieve-and-generate framework that first retrieves relevant knowledge from the knowledge base (KB) and then generates executable logical forms for answering questions. It focuses on leveraging retrieval to enhance logical form generation accuracy.

\textbf{DecAF}~\cite{DecAF} employs multi-granular retrieval strategies to ensure robust KBQA performance. By focusing on progressively refining retrieved knowledge, DecAF addresses the complexity of large-scale KBs and supports more accurate logical reasoning.

\textbf{TIARA}~\cite{TIARA} is a multi-stage retrieval method designed for large-scale KBs. It enhances robustness by retrieving candidate knowledge in multiple stages and integrating it into logical form generation, improving its ability to handle complex queries.

\textbf{SPARQA}~\cite{SPARQA} uses a skeleton-based semantic parsing approach to generate logical forms for complex questions. It simplifies question processing by extracting a structural skeleton, which is then converted into a complete logical form.

\textbf{BERT+Ranking}~\cite{GrailQA} integrates a BERT-based encoder with a ranking mechanism to select the most relevant answers from candidate entities. It enhances the precision of entity linking and relation matching in complex KBQA scenarios.

\textbf{ArcaneQA}~\cite{ArcaneQA} combines dynamic program generation with contextualized encoding to address complex reasoning tasks. Its innovative design enables it to process multi-hop reasoning queries with high accuracy.

The three low-resource baselines are designed to address the challenges of KBQA in scenarios. Unlike fully supervised methods, these approaches focus on GPT API for in-context-learning. Below is a detailed summary of each method:

\textbf{KB-BINDER}~\cite{KB-BINDER} leverages GPT-3.5-turbo to perform KBQA with minimal supervision. It adopts a structured in-context learning approach, where logical form templates guide the generation process. This method effectively balances efficiency and accuracy in low-resource settings.

\textbf{KB-Coder}~\cite{KB-Coder} employs code-style in-context learning to improve logical form generation. By treating logical form generation as a code-writing task, this method enhances reasoning consistency and adaptability, even with limited training data.

\textbf{ARG-KBQA}~\cite{ARG-KBQA} enhances knowledge retrieval efficiency in low-resource KBQA. It optimizes argument selection for logical forms, ensuring that queries retrieve the most relevant knowledge while minimizing errors caused by insufficient training data.

The four baselines in Section~\ref{Section5.4} represent different approaches to KBQA, covering both end-to-end and step-by-step methods. These methods are designed to evaluate different strategies, including retrieval-based logical form generation, sequential reasoning, and tree-based expansion. The comparison with KBQA-o1 highlights the strengths and limitations of each approach, providing insight into the effectiveness of heuristic exploration with MCTS. Below is a summary of each baseline:

\textbf{RG-E2E}: Based on the DecAF~\cite{DecAF}, RG-E2E follows a retrieve-then-generate paradigm, where relevant knowledge is first retrieved from the KB before generating the logical form. While effective in structured environments, this approach struggles with unseen entity-relation combinations due to its reliance on pre-retrieved data.

\textbf{GR-E2E}: Adapted from the ChatKBQA~\cite{ChatKBQA}, GR-E2E first generates a preliminary logical form and then refines it by retrieving supporting evidence from the KB. While more flexible than RG-E2E, it faces challenges in ensuring that the generated logical form aligns correctly with KB constraints.

\textbf{CoT-SbS}: Inspired by QueryAgent~\cite{QueryAgent}, CoT-SbS applies Chain-of-Thought (CoT) reasoning to KBQA, where logical forms are constructed incrementally through multiple reasoning steps. Although it improves interpretability, it is prone to local optima and reasoning errors, especially in complex multi-hop queries.

\textbf{ToT-SbS}: Based on the Think-on-Graph~\cite{ToG}, ToT-SbS extends step-by-step reasoning into a tree-like structure, expanding multiple reasoning paths simultaneously. While it mitigates local optima, it suffers from large search spaces, making it computationally expensive compared to heuristic search methods.

\section{Hyperparameter Settings}
\label{hyper}

Table~\ref{t_para} presents the hyperparameter configurations for the KBQA-o1 across three datasets: GrailQA, WebQSP, and GraphQ. These parameters are categorized into four stages: Initial Few-shot SFT, MCTS Exploration Stage, Incremental Fine-tuning, and MCTS Prediction Stage, each designed to optimize the KBQA framework’s performance for different tasks. 

In the Initial Few-shot SFT stage, the policy and reward models are fine-tuned using a small labeled dataset to initialize the framework. Both models adopt the DoRA architecture, optimized for reasoning tasks. The batch size is set to 4, ensuring stability in training with manageable memory usage. A fixed learning rate of 5e{-5} controls weight updates, providing a balance between convergence speed and training stability. The number of epochs varies by dataset, with 100 for the policy model in GrailQA and 50 for WebQSP and GraphQ. The reward model undergoes more training, with 300 epochs for GrailQA and 100 for WebQSP and GraphQ, to ensure the robustness of the reward function.

The MCTS Exploration Stage employs Monte Carlo Tree Search to explore logical forms by simulating reasoning paths within the KB. Each rollout performs 6 iterations of exploration. A beam size of 2 limits the number of candidate paths considered. TopK and TopD parameters further refine candidate selection, with GrailQA and GraphQ using TopK=10, while WebQSP uses TopK=3. Similarly, GrailQA and WebQSP have TopD=3, whereas GraphQ sets TopD=5. An exploration weight of 50 balances exploration of new paths with exploitation of known high-quality paths. The reward ratio, fixed at 0.5, ensures equal contribution from the policy and reward models. Dataset-specific reward thresholds (-100 for GrailQA, 30 for WebQSP, and -50 for GraphQ) filter out low-quality paths, tailoring the exploration process to the characteristics of each dataset.

The Incremental Fine-tuning stage further refines the policy and reward models based on insights gained during the exploration stage. The models retain the DoRA architecture, with a batch size of 4 and a learning rate of 5e{-5}. Both models are fine-tuned for 10 epochs across all datasets, ensuring improved generalization.

In the MCTS Prediction Stage, the refined models are used to generate final logical forms for KBQA tasks. 6 rollouts are performed, consistent with the exploration stage, but the beam size is reduced to 1 to focus on the most promising candidate paths. TopK and TopD parameters mirror those in the exploration stage to maintain consistency in candidate selection. The exploration weight is reduced to 10, prioritizing exploitation of high-quality paths while allowing limited exploration during prediction. The reward ratio remains fixed at 0.5 across all datasets, maintaining a consistent balance between contributions from the policy and reward models.

\begin{table}[h!t]
\centering
\caption{\label{t_para}
HyperParameter Settings for GrailQA, WebQSP, and GraphQ}
\vskip 0.1in
\small
\centering
\setlength{\tabcolsep}{5mm}{
\begin{tabular}{lccc}
\toprule
\textbf{Hyperparameter Name}          & \textbf{GrailQA} & \textbf{WebQSP} & \textbf{GraphQ} \\ \midrule
\multicolumn{4}{c}{\textit{Initial Few-shot SFT}} \\ \midrule
SFT1 Type for Policy Model       & DoRA             & DoRA            & DoRA            \\
SFT1 Batch Size for Policy Model & 4                & 4               & 4               \\
SFT1 Learning Rate for Policy Model & 5e-5          & 5e-5            & 5e-5            \\
SFT1 Epoch for Policy Model      & 100              & 50              & 50              \\
SFT1 Type for Reward Model       & DoRA             & DoRA            & DoRA            \\
SFT1 Batch Size for Reward Model & 4                & 4               & 4               \\
SFT1 Learning Rate for Reward Model & 5e-5          & 5e-5            & 5e-5            \\
SFT1 Epoch for Reward Model      & 300              & 100             & 100             \\ \midrule
\multicolumn{4}{c}{\textit{MCTS Exploration Stage}} \\ \midrule
MCTS1 Rollout $N$                 & 6                & 6               & 6               \\
MCTS1 Beam Size $B$                  & 2                & 2               & 2               \\
MCTS1 TopK $k$                       & 10               & 3               & 10              \\
MCTS1 TopD $d$                       & 3                & 3               & 5               \\
MCTS1 Exploration Weight $w$         & 50               & 50              & 50              \\
MCTS1 Reward Ratio $\delta$               & 0.5              & 0.5             & 0.5             \\
MCTS1 Reward Threshold $\gamma^*$           & -100             & 30              & -50             \\ \midrule
\multicolumn{4}{c}{\textit{Incremental Fine-tuning}} \\ \midrule
SFT2 Type for Policy Model       & DoRA             & DoRA            & DoRA            \\
SFT2 Batch Size for Policy Model & 4                & 4               & 4               \\
SFT2 Learning Rate for Policy Model & 5e-5          & 5e-5            & 5e-5            \\
SFT2 Epoch for Policy Model      & 10               & 10              & 10              \\
SFT2 Type for Reward Model       & DoRA             & DoRA            & DoRA            \\
SFT2 Batch Size for Reward Model & 4                & 4               & 4               \\
SFT2 Learning Rate for Reward Model & 5e-5          & 5e-5            & 5e-5            \\
SFT2 Epoch for Reward Model      & 20               & 20              & 20              \\ \midrule
\multicolumn{4}{c}{\textit{MCTS Prediction Stage}} \\ \midrule
MCTS2 Rollout $N$                    & 6                & 6               & 6               \\
MCTS2 Beam Size $B$                  & 1                & 1               & 1               \\
MCTS2 TopK $k$                       & 10               & 3               & 10              \\
MCTS2 TopD $d$                       & 3                & 3               & 5               \\
MCTS2 Exploration Weight $w$         & 10               & 10              & 10              \\
MCTS2 Reward Ratio $\delta$               & 0.5              & 0.5             & 0.5             \\ \bottomrule
\end{tabular}}
\end{table}

\section{Open-source LLMs used in KBQA-o1}
\label{play}
The KBQA-o1 framework is designed with a plug-and-play architecture, allowing seamless integration of different open-source LLMs based on task requirements. This flexibility allows easy substitution or upgrading of LLMs, ensuring that the system remains state-of-the-art as newer models become available.

\textbf{Llama-3}~\cite{llama3}, developed by Meta AI, is an advanced large language model known for its strong reasoning abilities and generalization across diverse tasks. It excels in processing multi-hop reasoning and compositional queries, making it ideal for KBQA tasks that require logical inference over structured data, ensuring robustness in complex question-answering scenarios.

\textbf{Qwen2.5}~\cite{Qwen2.5}, released by Alibaba, is particularly effective in entity linking and relation extraction, two crucial components of KBQA. Trained on both structured and unstructured data, Qwen2.5 ensures high-precision entity disambiguation and improves the accuracy of knowledge retrieval. Its multilingual capabilities make it promising to further enhance KBQA-o1’s adaptability across different datasets and domains.

\textbf{Gemma-2}~\cite{Gemma2}, developed by Google, is optimized for retrieval-based reasoning and structured query generation. It specializes in efficiently retrieving relevant knowledge and generating accurate logical forms for complex queries. Its lightweight yet powerful design ensures high-speed inference, making it well-suited for low-latency applications in KBQA.

\section{Case Study}

\begin{figure}[h]
\centering
\includegraphics[width=17cm]{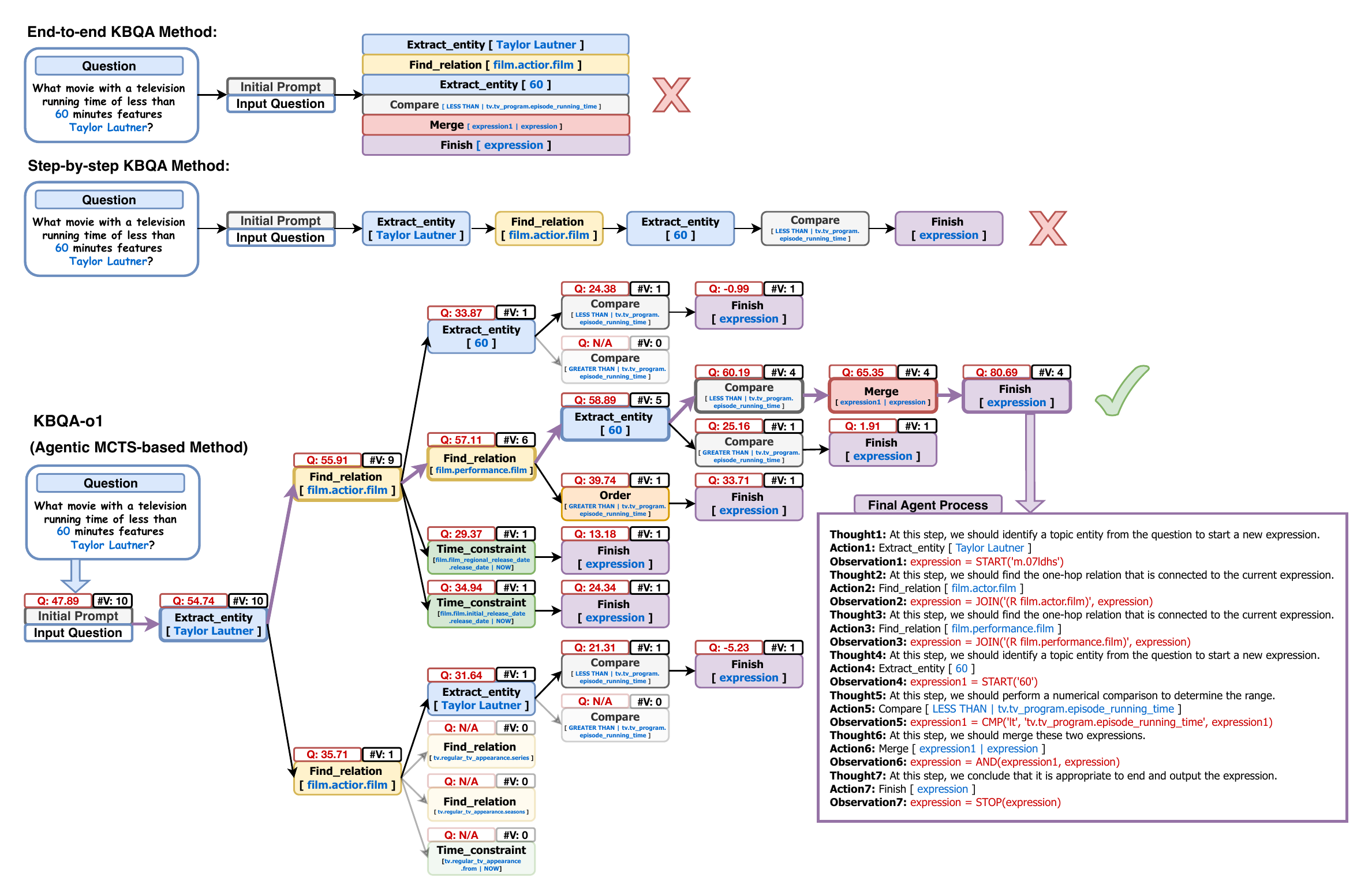}
\caption{Case study of KBQA-o1's MCTS inference compared with end-to-end and step-by-step methods.}
\label{f_case}
\end{figure}

To visually illustrate the reasoning process of KBQA-o1, Figure~\ref{f_case} compares the proposed MCTS-based method with end-to-end and step-by-step KBQA approaches. In this example, the question “What movie with a television running time of less than 60 minutes features Taylor Lautner?” is posed. The question is first processed in the initial prompt, forming the agent’s starting state. Through multiple rounds of MCTS exploration, the framework builds a search tree, where the Q-value represents the score updated through backpropagation, and \#V denotes the number of times each state has been visited. These values guide the agent in selecting the optimal next node using the Upper Confidence Bound for Trees (UCT).

The end-to-end KBQA method prematurely terminates logical form generation due to incomplete reasoning and a lack of exploration. The step-by-step KBQA method incrementally constructs logical forms but also fails, as it cannot fully navigate the reasoning constraints, leading to an incorrect result. In contrast, the KBQA-o1 method effectively explores and evaluates multiple reasoning paths, refining its decisions based on reward feedback and generating the correct logical form. The final agent process demonstrates how observations at each step are transformed into graph query statements, which are executed to obtain the correct answer. This process highlights KBQA-o1’s ability to systematically decompose the query, balance exploration and exploitation, and handle multi-step reasoning effectively, outperforming other methods.

\section{Error Analysis}
We analyze the errors made by KBQA-o1 on the GrailQA dataset and categorize them into three main types. These errors reflect the challenges in exploring logical paths, retrieving accurate results, and ranking the most appropriate paths in the context of complex KBQA tasks.

\textbf{Executable path not discovered (29.8\%)} In this category, KBQA-o1 fails to explore any executable logical path in the knowledge base, often due to the complexity of GrailQA’s compositional and zero-shot queries. These errors arise from insufficient exploration of multi-hop reasoning paths and limited representation of rare logical forms in the training data. Improvements in exploration depth and diversity of logical forms in the training process could help mitigate this issue.

\textbf{Correct path not discovered (54.7\%)} This is the most common error type, where KBQA-o1 explores logical paths but fails to discover the correct one. These errors highlight challenges in precise entity and relation retrieval, particularly in zero-shot settings where entities or relations are unseen during training. Enhancing the semantic understanding and retrieval mechanisms would be crucial for reducing this type of error.

\textbf{Correct path not selected as the best (15.5\%)} In some cases, KBQA-o1 successfully explores the correct logical path but fails to select it as the optimal solution. For example, it may assign a higher Q-value to an incorrect logical form over the correct one. These errors often stem from suboptimal reward evaluation during MCTS backpropagation or inconsistencies in the reward model’s fine-tuning. Addressing these issues through better reward modeling and fine-tuning could significantly enhance path selection accuracy.

\section{Future Directions}
\label{future}
The KBQA-o1 offers several promising directions for future research and development. These directions aim to enhance the model’s capabilities, extend its application domains, and address current limitations.

\textbf{Exploring Reinforcement Learning Techniques Like DPO for Continual Learning.} One promising direction is leveraging advanced reinforcement learning techniques, such as Direct Policy Optimization (DPO), to enable continual learning for KBQA-o1. By refining policy and reward models through reinforcement learning, the framework could dynamically adapt to new datasets and tasks. A key challenge in this direction is constructing high-quality positive and negative sample pairs to guide the learning process effectively. Future work could explore automated methods for generating these samples or employing human-in-the-loop systems for quality assurance.

\textbf{Expanding the Set of Logical Operators for Enriched QA Patterns.} To broaden the range of questions the system can handle, future research could focus on introducing and supporting more diverse logical operators. This would allow KBQA-o1 to answer more complex queries, particularly those requiring advanced reasoning patterns such as conditional logic, aggregation, and temporal constraints. By integrating these operators, the framework could offer more comprehensive and flexible question-answering capabilities, covering a wider array of user needs.

\textbf{Specialized Applications in Domains like Medicine and Law.} KBQA-o1 has significant potential for domain-specific applications, particularly in fields such as medicine and law. These areas require precise reasoning and domain-specific knowledge to handle intricate and high-stakes queries. Adapting the framework to these domains would involve integrating specialized knowledge bases and fine-tuning the models to align with domain-specific terminologies and reasoning patterns. Such advancements could pave the way for impactful real-world applications, including clinical decision support and legal research.

\textbf{Extending to Multimodal, Multilingual, and Multi-Agent Systems.} Future iterations of KBQA-o1 could extend its capabilities to multimodal and multilingual settings, enabling the framework to process and reason over diverse input formats such as images, audio, and text in multiple languages. Additionally, incorporating multi-agent systems could enhance the framework’s ability to handle collaborative and interactive tasks, where multiple agents contribute to exploring and resolving complex queries. These advancements would significantly expand the framework’s usability across various scenarios.


\end{document}